%% file: paper.tex
\documentclass{article}
\usepackage{ijcai24}

\usepackage{times}
\usepackage{soul}
\usepackage{url}
\usepackage[hidelinks]{hyperref}
\usepackage[utf8]{inputenc}
\usepackage[small]{caption}
\usepackage{graphicx}
\usepackage{amsmath}
\usepackage{amsthm}
\usepackage{booktabs}
\usepackage{algorithm}
\usepackage[switch]{lineno}
\usepackage{subcaption}
\usepackage[flushleft]{threeparttable}
\usepackage{amsfonts, xspace, xcolor}
\usepackage{algorithmicx, algpseudocode}

\include{define}

\title{Enhancing Molecular Property Prediction with Auxiliary Learning and Task-Specific Adaptation}

\author{
Vishal Dey$^1$
\and
Xia Ning$^1$
\affiliations
$^1$Computer Science and Engineering, The Ohio State University\\
\emails
\{dey.78, ning.104\}@osu.edu
}

\begin{document}

\maketitle

\begin{abstract}
Pretrained Graph Neural Networks have been widely adopted for various
molecular property prediction tasks.
Despite their ability to encode structural and relational features of molecules,
traditional fine-tuning of such pretrained GNNs on the target task can lead to poor generalization.
To address this, we explore
the adaptation of pretrained GNNs to the target task by jointly training them with multiple auxiliary tasks.
This could enable the GNNs to learn both general and task-specific features, which may benefit the target task.
%
However, a major challenge is to determine
the relatedness of auxiliary tasks with the target task.
%
To address this, we investigate multiple strategies to
measure the relevance of auxiliary tasks and integrate such tasks 
by adaptively combining task gradients
%
or by learning task weights via bi-level optimization.
Additionally, we {propose} a novel gradient surgery-based approach,
{Rotation of Conflicting Gradients (\RCGrad),}
that learns to align conflicting auxiliary task gradients through rotation.
%
Our experiments with state-of-the-art pretrained GNNs demonstrate the efficacy of our proposed methods,
with improvements of up to {7.7}\% over fine-tuning.
%
This suggests that incorporating auxiliary tasks
along with target task fine-tuning
can be an effective way to improve the generalizability of pretrained GNNs
for molecular property prediction.
%
\end{abstract}


\section{Introduction}
\label{sec:intro}

Accurate prediction of molecular properties is pivotal in drug discovery~\cite{wieder2020compact},
as it accelerates the identification of potential molecules with desired properties.
Developing computational models for property prediction relies on learning effective representations of molecules~\cite{david2020molecular}.
In this regard, Graph Neural Networks (GNNs) have shown impressive results
in learning effective representations for molecular property prediction tasks~\cite{gasteiger2021directional,wang2022molecular,guo2023graph}.
Inspired by the paradigm of pretraining followed by fine-tuning,
widely recognized for its impact in natural language understanding~\cite{radford2018improving,wei2022emergent}, 
molecular GNNs are often pretrained~\cite{hu2019strategies} on a large corpus of molecules.
%
Such a corpus might encompass irrelevant data for the target property prediction task. 
This can lead the GNNs to learn features that do not benefit the target task.
Consequently, pretrained GNNs are fine-tuned with the target task to encode task-specific features. 
However, vanilla fine-tuning can potentially lead to poor generalization,
particularly when dealing with diverse downstream tasks, 
limited data, and the need to generalize across varying scaffolds~\cite{wu2018moleculenet}. 

To improve generalization, auxiliary learning has recently garnered attention~\cite{liebel2018auxiliary,liu2019self,dery2022aang}.
Auxiliary learning leverages informative signals from self-supervised tasks on unlabeled data,
to improve the performance of the target tasks. 
However, its application in the context of molecular graphs, 
specifically for molecular property prediction, remains largely unexplored.
Following this line of work, {in this paper,}
we explore how to adapt pretrained molecular GNNs by combining widely-used self-supervised 
tasks with the target task 
using respective {task-specific data (with self-supervised and target task labels)}.
However, a critical challenge in such an adaptation is caused by negative transfer\cite{rosenstein2005transfer},
where auxiliary tasks might impede rather than aid the target task~\cite{ruder2017overview,du2018adapting}.

{To address this challenge}, we develop novel gradient surgery-based adaptation strategies, referred to as
Rotation of Conflicting Gradients (\RCGrad) and
Bi-level Optimization with Gradient Rotation (\BLORC).
Such strategies mitigate negative transfer from auxiliary tasks by learning to align conflicting gradients.
Overall, our adaptation strategies improved
the target task performance by as much as {7.7}\% over vanilla fine-tuning.
Moreover, our findings indicate that the 
developed adaptation strategies are particularly effective in tasks with limited labeled data,
which is a common challenge in molecular property prediction tasks.
Our comprehensive investigation of multiple adaptation strategies for pretrained molecular GNNs
represents a notable contribution
in addressing the limited benefit of pretrained GNNs~\cite{sun2022does},
and in improving generalizability across a diverse set of downstream tasks with limited data.

\section{Related Work}
\label{sec:app:rel}

\subsection{Pretraining and fine-tuning GNNs}
\label{sec:app:rel:ft}
Pretraining followed by fine-tuning is widely used to leverage knowledge gained from
related tasks and to improve model generalization.
Typically, it involves training a model on large-scale data with self-supervised 
or supervised tasks, and then fine-tuning it on a small-scale labeled data.
{Following the success of pretraining and fine-tuning paradigm in various domains}~\cite{liu2019roberta,floridi2020gpt},
researchers have extended it to molecular GNNs~\cite{hu2019strategies,hu2020gpt,liu2021pre,wang2022molecular}.
In this regard, researchers have designed a number of self-supervised tasks as pretraining tasks
that focus on capturing diverse chemical rules, connectivities, and patterns at
varying granularities: on node, subgraph and graph levels~\cite{xia2022pre}.
%
Although pretrained GNNs showed promise in capturing diverse chemical knowledge,
the challenge lies in effectively extracting this knowledge relevant to the target task,
which is often non-trivial through vanilla fine-tuning.
Specifically, such fine-tuning often leads to overfitting~\cite{xia2022towards}.
Contrary to the observations in domains such as natural language processing (NLP) and computer vision,
where pretrained models consistently yield substantial improvements,
pretrained GNNs do not exhibit such improvement~\cite{sun2022does}.

This could be due to a notable research gap in determining what self-supervised molecular tasks
can better benefit the downstream target tasks.
In fact, prior studies in pretraining molecular GNNs mostly leverage one or two
self-supervised task(s), thereby resulting in a plethora of multiple pretrained GNNs.
Interestingly, such pretrained GNNs capture different knowledge~\cite{wang2022evaluating}
and excel in different downstream
molecular property prediction tasks~\cite{sun2022does}.
Additionally, Sun et al.~\cite{sun2022does} recently demonstrated that
self-supervised graph pretraining does not consistently/significantly outperform
non-pretraining methods across various settings.
Overall, although pretrained GNNs hold promise for molecular property prediction,
their benefit over non-pretrained models seems limited. 
{To address this, some recent attempts~\cite{xia2022towards,zhang2022fine}
to fine-tune pretrained GNNs have largely relied on existing ideas like regularization~\cite{xuhong2018explicit}
or update constraints~\cite{houlsby19parameter} during fine-tuning.
In contrast, our proposed approaches leverage auxiliary tasks
to learn generalizable knowledge and prevent overfitting to the training set.}

\subsection{Knowledge Transfer with Auxiliary Learning}
\label{sec:app:rel:aux}
Knowledge transfer through auxiliary learning
has demonstrated its effectiveness across a spectrum of domains
~\cite{trinh2018learning,nediyanchath2020multi,lee2021improving}.
This paradigm, distinct from multi-task learning,
aims to optimize the target task's performance while leveraging auxiliary tasks to bolster generalization~\cite{shi2020auxiliary}. 
Prior research in other domains has developed multiple methods to automatically
learn task weights, such as using gradient similarity~\cite{dery2021auxiliary,du2018adapting},
using parameterized auxiliary network~\cite{navon2020auxiliary,dery2022aang},
using bi-level optimization and implicit differentiation~\cite{navon2020auxiliary,pmlr-v162-chen22y},
minimizing distances between task embeddings~\cite{chen2021weighted},
or from the perspective of Nash equilibrium~\cite{shamsian2023auxiliary}.
However, the application of auxiliary learning for adapting molecular 
GNNs to target tasks, particularly in the context of molecular property prediction, remains an under-explored area.
In this study, we adopt and explore gradient similarity, gradient scaling, and bi-level optimization strategies.

\section{Preliminaries}
\label{sec:prelim}

Motivated by the success of continued pretraining and task-specific adaptation in pretrained Large Language Models (LLMs)
~\cite{gururangan2020don,dery2021should,yang2022c3},
we investigate adaptation of off-the-shelf pretrained molecular GNNs
to target molecular property prediction tasks. 
Via such an adaptation,
we aim to leverage existing self-supervised (SSL) tasks designed for molecular GNNs
and transfer learned knowledge from such tasks to the target task.
%
%
We employ the existing SSL tasks typically used in molecular pretraining  
such as
masked atom prediction (AM),
context prediction (CP)~\cite{hu2019strategies}, edge prediction (EP)~\cite{Hamilton:2017tp}, 
graph infomax (IG)~\cite{sun2019infograph}, 
and motif prediction (MP)~\cite{rong2020self}.
We refer to these tasks as auxiliary tasks.
%
Intuitively, these auxiliary tasks can potentially capture diverse chemical semantics and rich
structural patterns at varying granularities.
By utilizing SSL objectives on target task-specific data,
auxiliary tasks augment the pretrained GNNs with richer representations.
Such representations, in turn, can improve the generalizability of the target property prediction task.
Henceforth, the term ``GNN'' refers to an off-the-shelf pretrained molecular GNN.

\begin{figure}[t!]
    \centering
    \begin{subfigure}{0.5\textwidth}
    \centering
        \includegraphics[scale=0.5]{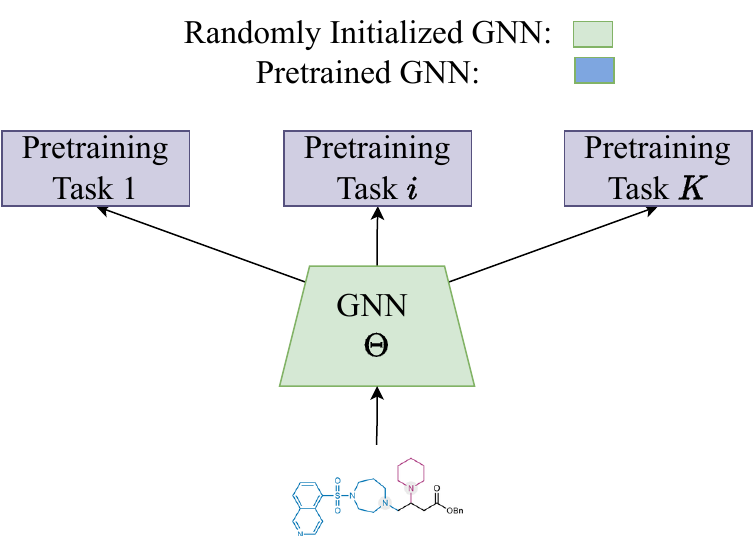}
        \caption{Pretraining Stage}
    \end{subfigure}
    \hfill
    \begin{subfigure}{0.5\textwidth}
    \centering
        \includegraphics[scale=0.5]{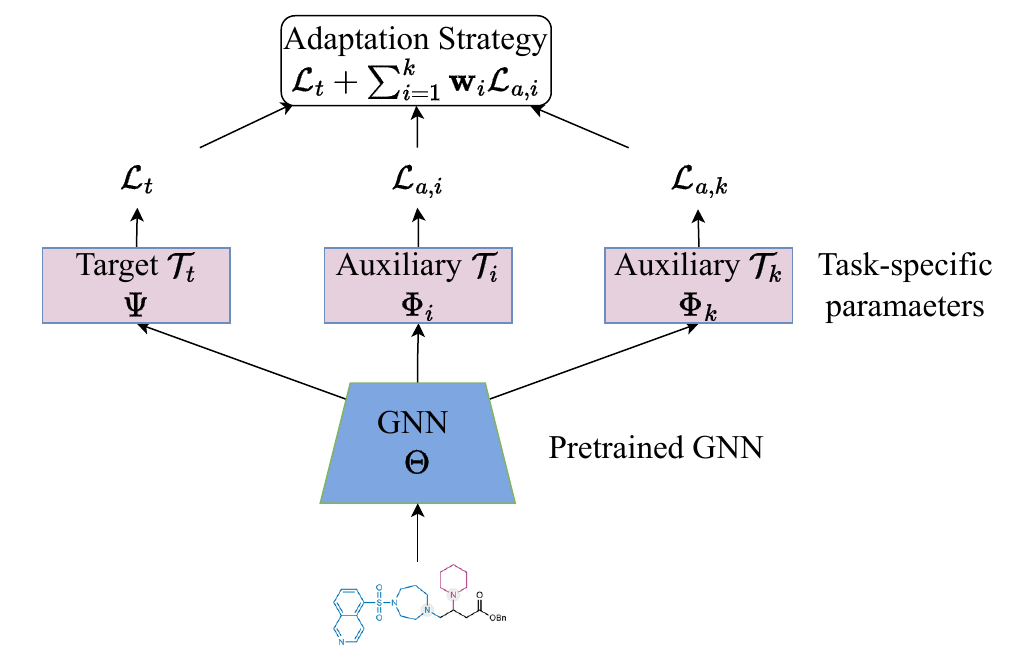}
        \caption{Adaptation Stage}
    \end{subfigure}
     \caption{Off-the-shelf available pretrained GNNs are
     transferred for target task-specific adaptation.}
    \label{fig:model}
\end{figure}
Figure \ref{fig:model} presents an overview of the adaptation setup.
Formally, we adapt a GNN with parameters \ParGnn
to optimize the performance on the target task \TgtTask.
We achieve this by
jointly training \TgtTask with auxiliary tasks $\{\iAuxTask\}^k_{i=1}$
{through solving the following optimization problem}:
\begin{equation}
\label{equ:loss}
\min_{\scriptsize{\ParGnn,\ParTgt,\ParAux}_{i\in\{1..k\}}} \TgtLoss + 
\sum_{i=1}^k\TaskWeight_i\iAuxLoss,
\end{equation}
where \TgtLoss and \iAuxLoss denote the target task loss and $i$-th auxiliary task loss, respectively,
\ParTgt and $\ParAux_{i\in\{1,...,k\}}$ denotes task-specific learnable parameters
for the target and $i$-th auxiliary task, respectively,
and $\TaskWeight$ is the weight indicating the influence of the auxiliary tasks on the target task.
Through the above optimization,
all the parameters are simultaneously updated in an end-to-end manner.
Note that the above optimization does not optimize \TaskWeight -- we will introduce an approach that
can additionally learn \TaskWeight {in Section~\ref{sec:methods:blo}}.
In fact, the key to effective adaptation lies in accurately determining $\TaskWeight$,
such that the combined task gradients can backpropagate relevant training signals to the shared GNN as follows:
\vspace{-1pt}
\[\ParGnn^{(t+1)} := \ParGnn^{(t)} - \alpha
\left(\TgtGrad + \sum\nolimits_{i=1}^k \TaskWeight_i \iAuxGrad\right),\]
\vspace{-1pt}
where $\TgtGrad = \GradwParGnn\TgtLoss$, and $\iAuxGrad = \GradwParGnn\iAuxLoss$
denote the gradients updating \ParGnn from the target and $i$-th auxiliary task, respectively,
and $\alpha$ denotes the learning rate.
%
%
Our proposed adaptation strategies focus on learning such \TaskWeight in an end-to-end manner,
to dynamically combine task gradients during each update.
These strategies contrast with those using fixed weights or conducting expensive grid-search to explore all possible \TaskWeight.
%

\subsection{Gradient Cosine Similarity (\GradSim)}
%
The first strategy to meaningfully combine task gradients is based on gradient cosine similarity (\GradSim)~\cite{du2018adapting}.
Intuitively, \GradSim measures the alignment between task gradients during training,
providing insights into the relatedness of auxiliary tasks with the target task. 
A high \GradSim indicates that the auxiliary tasks provide complementary information,
and thus, can benefit the target task.
Conversely, low \GradSim indicates potential orthogonality or even conflict between tasks.
Thus, \GradSim can naturally quantify the relatedness of auxiliary tasks 
with the target task over the course of training.
%
We compute \GradSim and update \ParGnn as:
\vspace{-5pt}
\[\ParGnn^{(t+1)} := \ParGnn^{(t)} - \alpha
\left(\TgtGrad + \sum\nolimits_{i=1}^k \max\left(0, \cos\left(\TgtGrad, \iAuxGrad)\right)
\iAuxGrad\right)\right),\]
where, $\max$ operator takes the maximum out of the two values, thereby,
dropping the tasks with conflicting gradients (i.e., with negative \GradSim).

\begin{figure}[!h]
    \centering
    \begin{subfigure}{0.5\textwidth}
     \centering
    \includegraphics[scale=0.08]{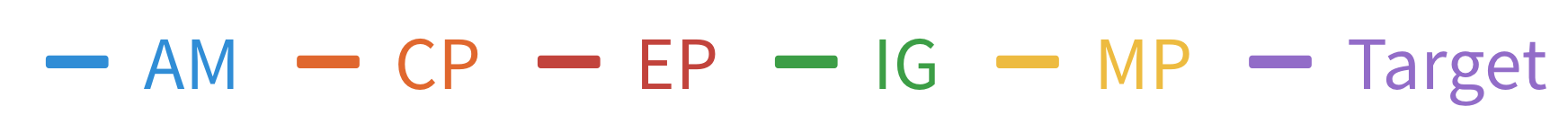}
    \end{subfigure}
        \begin{subfigure}{0.15\textwidth}
        \includegraphics[width=\textwidth]{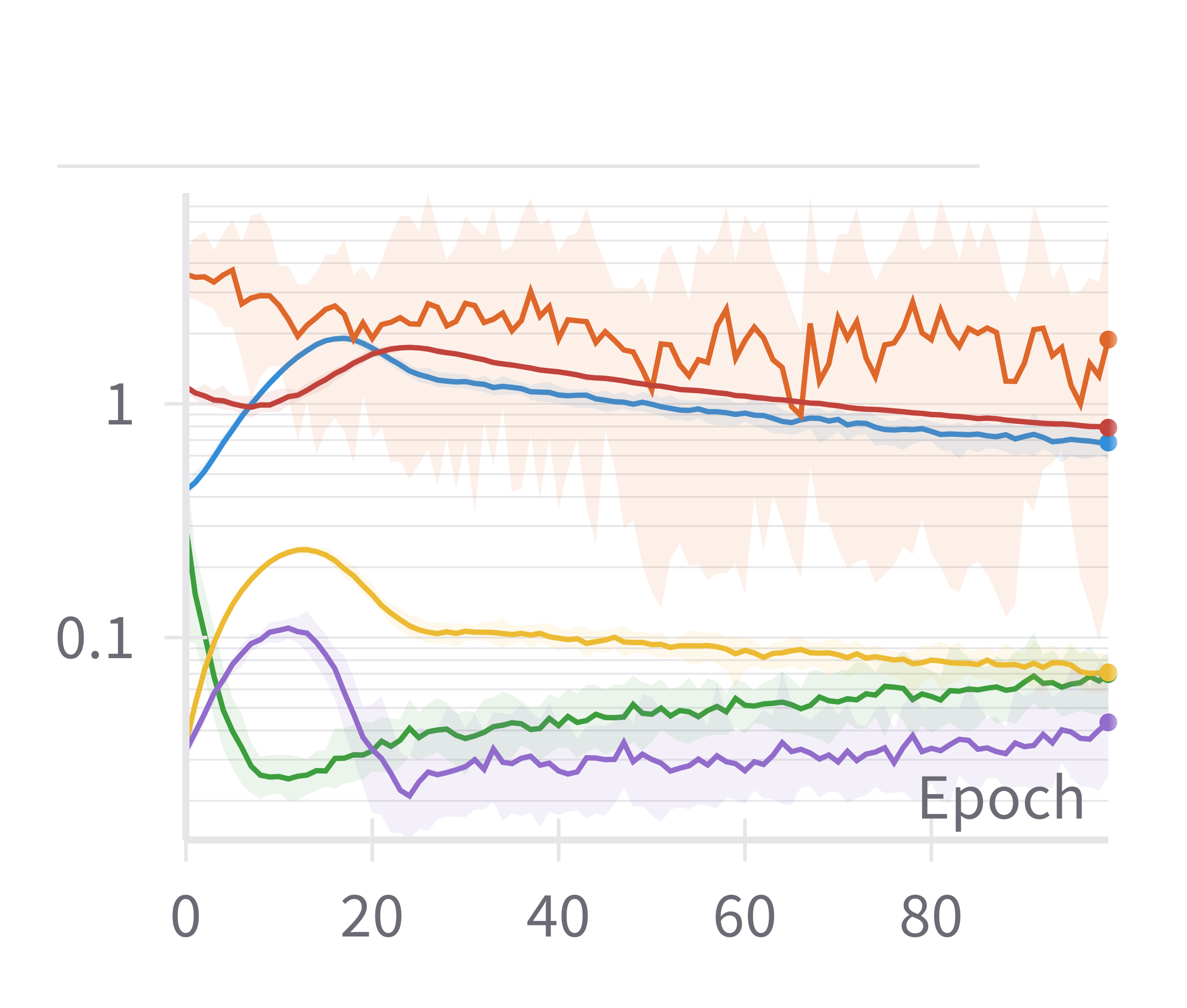}
        \caption{SIDER}
    \end{subfigure}
    \begin{subfigure}{0.15\textwidth}
        \includegraphics[width=\textwidth]{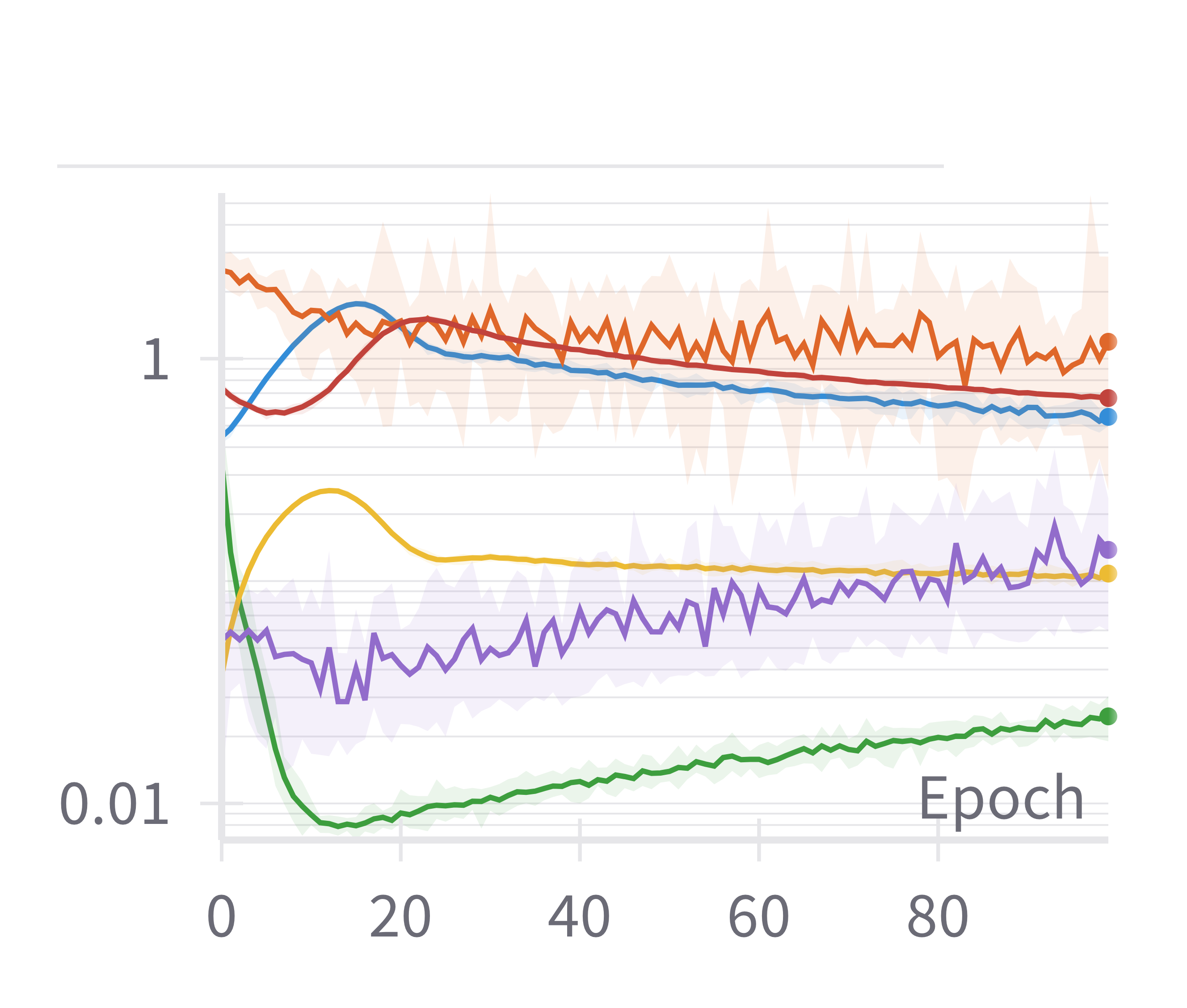}
        \caption{BACE}
    \end{subfigure}
     \begin{subfigure}{0.15\textwidth}
        \includegraphics[width=\textwidth]{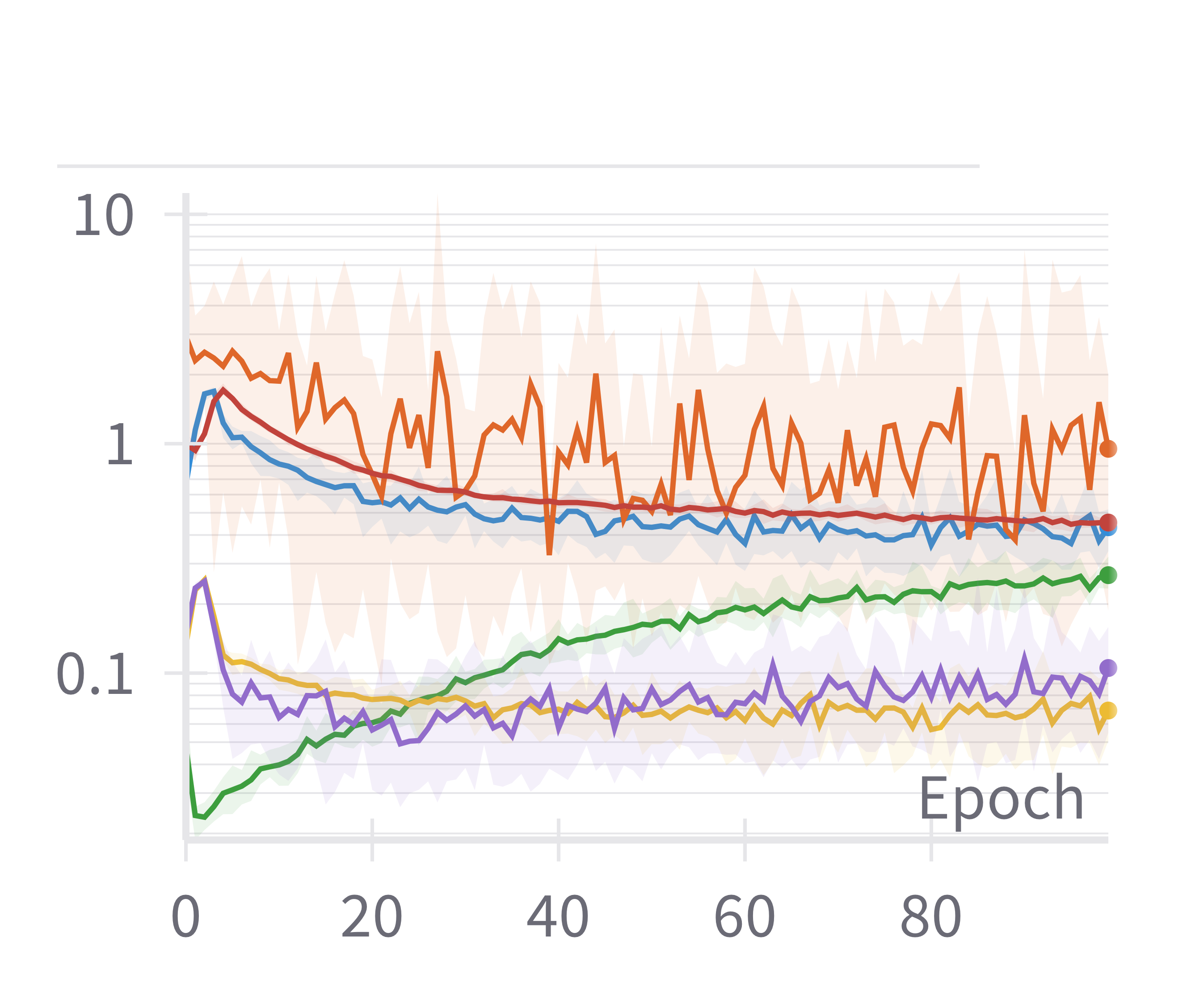}
        \caption{Tox21}
    \end{subfigure}    
    \caption{Large variations of scales among task gradients are observed when
    \SUPC is adapted with all auxiliary tasks using \MTL.}
    \label{fig:supc:gnorm}
\end{figure}
%

\subsection{Gradient Scaling (\GradScale)}
We also adopt a simpler strategy of gradient scaling~\cite{he2022metabalance}
to adjust the influence of
auxiliary tasks with respect to the target task.
Our preliminary experiments as presented in Figure~\ref{fig:supc:gnorm}
revealed significant differences in the scales of
the task gradient norms, and thus requiring careful adjustments.
This is because if the gradient of an auxiliary task is much larger than that of the target task,
\ParGnn updates will be most dominated by such auxiliary tasks,
thereby potentially resulting in worse target performance.
On the other hand, if the gradient of an auxiliary task is relatively small,
the training signals from such auxiliary tasks will be too weak to encode
any relevant features in \ParGnn.
Thus, following~\cite{chen2018gradnorm,he2022metabalance}, we use a simple gradient scaling to dynamically adjust
the influence of auxiliary tasks during updates of \ParGnn as follows:
\begin{equation}
\label{equ:gns}
\!\!\!\!\ParGnn^{(t+1)} := \ParGnn^{(t)} - \alpha
\left(\TgtGrad + \sum_{i=1}^k \max\left(1, \frac{||\TgtGrad||}{||\iAuxGrad||}\right) \iAuxGrad\right),
\end{equation}
where $||\cdot||$ denotes the {$\ell$-2} norm.

\section{Methods}
\label{sec:methods}
    
\begin{figure*}
\centering
	\begin{subfigure}{\textwidth}
	\centering
	\includegraphics[width=\textwidth]{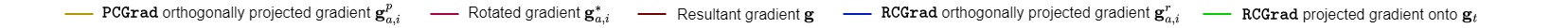}
	 \end{subfigure}
    \begin{subfigure}{0.23\textwidth}
        \includegraphics[width=1.2\textwidth]{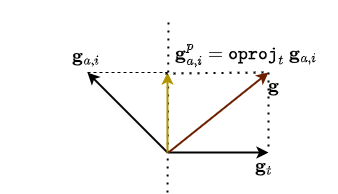}
        \caption{\PCGrad}
        \label{fig:pcgrad}
    \end{subfigure}
    \hfill
    \begin{subfigure}{0.26\textwidth}
    \centering
        \includegraphics[width=1.2\textwidth]{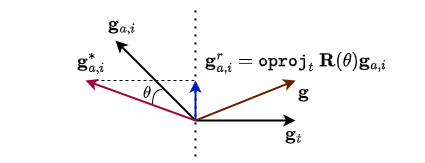}
        \caption{\RCGrad}
        \label{fig:rcgradsub}
    \end{subfigure}
    \hfill
      \begin{subfigure}{0.25\textwidth}
      \centering
        \includegraphics[width=1.2\textwidth]{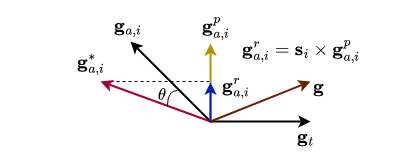}
         \caption{Scaling \iAuxGradP by $\RWeight_i$}
         \label{fig:scalingS}
    \end{subfigure}
    \hfill
     \begin{subfigure}{0.24\textwidth}
     \centering
        \includegraphics[width=1.2\textwidth]{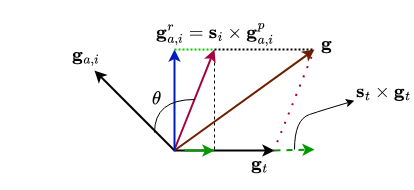}
         \caption{Scaling \iAuxGradP and \TgtGrad}
         \label{fig:scalingG}
    \end{subfigure}

        \caption{(a) \PCGrad projects conflicting gradient \iAuxGrad onto the normal plane of \TgtGrad.
        (b) \RCGrad applies a rotation to \iAuxGrad, followed by projection.
        (c) Rotation followed by orthogonal projection is equivalent to scaling \iAuxGradP.
        (d) If the rotated gradient does not conflict with \TgtGrad, 
	the projection of the rotated gradient onto \TgtGrad is incorporated as scaling \TgtGrad by $(1+\RWeight_t)$.}
        \label{fig:rcgrad}
\end{figure*}

\subsection{Rotation of Conflicting Gradients (\RCGrad)}
\label{sec:methods:rcgrad}
%
While both conflicting directions and magnitude differences of task gradients can lead to negative transfer,
\GradSim and \GradScale focus separately on homogenizing either the direction or magnitude of gradients, rather than in a unified manner.
To address these limitations,
we develop Rotation of Conflicting Gradients (\RCGrad) --
a novel extension of \PCGrad~\cite{yu2020gradient} -- 
that aligns gradients both in terms of direction and magnitude.
\RCGrad, which builds upon \PCGrad,
does not completely discard gradients conflicting with the target task, unlike \GradSim.
Instead, \RCGrad only negates the component of the conflicting gradient that
is completely opposite to the target task gradient.
Additionally, \RCGrad explicitly learns how much of the non-conflicting component should be incorporated
for the most effective knowledge transfer.
This mitigates negative transfer by not only removing the conflicting component but also by learning to incorporate 
a portion of the non-conflicting component.

Figure~\ref{fig:rcgrad} demonstrates the difference between \PCGrad and \RCGrad.
Formally, \RCGrad learns to rotate auxiliary gradient $\mathbf{g}_{a,i}$ by angle $\theta_i$
to yield a rotated gradient $\mathbf{R}(\theta_i)\iAuxGrad$, which is followed by an orthogonal projection 
in case of conflicts (Figure~\ref{fig:rcgradsub}). 
The orthogonally projected component is computed as 
$\iAuxGradR =  \mathtt{oproj}_{t} ~\mathbf{R}(\theta_i)\iAuxGrad$,
where $\mathbf{R}(\theta_i)$ is the rotation matrix parameterized by $\theta_i$,
and $\mathtt{oproj}_t$ is the orthogonal vector projection operator
as defined in Equation~\ref{equ:pcgrad}.
Via such an operator (Figure~\ref{fig:pcgrad}), 
\PCGrad projects the conflicting auxiliary gradient \iAuxGrad onto the normal
plane of the target task's gradient \TgtGrad to yield $\iAuxGradP$ as follows:
\begin{equation}
\label{equ:pcgrad}
\iAuxGradP = \mathtt{oproj}_{t} ~\iAuxGrad
= \iAuxGrad - \frac{\iAuxGrad \cdot \TgtGrad}{||\TgtGrad||} \cdot \frac{\TgtGrad}{||\TgtGrad||},
\end{equation}
where $\mathtt{oproj}_t$ denotes the orthogonal projection operator with respect to \TgtGrad.
This enables effective knowledge transfer from auxiliary tasks, even if they share some dissimilarity to the target task.
However, \PCGrad does not explicitly learn how much of the non-conflicting component should be incorporated
for the most effective knowledge transfer.
To address this limitation, \RCGrad learns an appropriate rotation to be applied to the auxiliary gradient \iAuxGrad,
followed by the projection of the rotated gradient.
Such a learnable rotation in an end-to-end manner enables dynamic knowledge transfer from auxiliary tasks such that the
target task performance can be improved.
%
%

%
%
%
Moreover, as shown in Figures~\ref{fig:scalingS} and \ref{fig:scalingG},
the rotation followed by the projection of gradients is equivalent to applying
appropriate scaling factors $\RWeight_i$ and $\RWeight_t$ 
on the projected gradients \iAuxGradP and \TgtGrad, respectively.
Additionally, different from \PCGrad, \RCGrad accounts for large differences in gradient magnitudes
by adjusting the magnitudes of non-conflicting auxiliary task gradients relative to that of the target task gradient
(Equation~\ref{equ:gns}).
To summarize, \ParGnn is updated as follows:
$\ParGnn^{(t+1)} := \ParGnn^{(t)} - \alpha \mathbf{g}$, where 
\begin{equation}
\label{equ:rcgrad}
\mathbf{g} = 
\begin{cases}
(1+\RWeight_t) \times \TgtGrad +\sum_{i=1}^k \RWeight_i \times \iAuxGradP,  &  \text{if $\TgtGrad \cdot \iAuxGrad < 0$}\\
\TgtGrad + \sum_{i=1}^k \max\left(1, \frac{||\TgtGrad||}{||\iAuxGrad||}\right) \iAuxGrad,  & \text{otherwise}
\end{cases}
\end{equation}
where \iAuxGradP is computed via equation~\ref{equ:pcgrad}.
Note that the set of scaling factors $\RWeight=\{\{\RWeight_i\}_{i=1}^k, \RWeight_t\}$ 
is learned in an end-to-end manner during the optimization of the 
combined losses from all tasks.

\subsection{Bi-Level Optimization (\BLO)}
\label{sec:methods:blo}
Unlike the previous approaches that directly manipulate task gradients, 
\BLO learns task weights \TaskWeight (Equation~\ref{equ:loss})
in an end-to-end manner,
such that the GNN generalizes well to the target task.
Note that \BLO does not directly intervene in the gradient computation process. 
Instead, \BLO learns \TaskWeight that minimizes the target validation loss
while ensuring that the GNN 
is optimized with a weighted combination of losses:
\begin{equation}
\begin{aligned}
\label{equ:biopt}
\TaskWeight^* &= \arg\min\nolimits_{\scriptsize{\TaskWeight}}
\TgtLoss^{\scriptsize{(\AuxData)}}(\ParGnn^*(\TaskWeight)),  \\
\text{s.t.} ~~~~~~~
\ParGnn^*(\TaskWeight) &= \arg\min\nolimits_{\scriptsize{\ParGnn}} \TotalLoss(\ParGnn, \TaskWeight)
\end{aligned}
\end{equation}
where, $\TotalLoss = \TgtLoss + \sum_{i=1}^k\TaskWeight_i\iAuxLoss$ is the
combined loss on the training set, and
$\TgtLoss^{\scriptsize{(\AuxData)}}$ is the loss on the target task
computed with a held-out auxiliary dataset \AuxData,
and $\ParGnn^*(\TaskWeight)$ is the best-response of \ParGnn with current \TaskWeight.
This formulation is a bi-level optimization problem:
updating \TaskWeight in the upper-level optimization requires computing
$\GradwTaskWeight\TgtLossOnAuxD = \GradwParGnn\TgtLossOnAuxD \cdot \GradwTaskWeight \ParGnn^*$,
where the latter gradient requires back-propagation through the inner-level
optimization of \ParGnn.
Following ~\cite{lorraine2020optimizing}, we leverage the Implicit Function Theorem (IFT)
to compute 
$\GradwTaskWeight \ParGnn^* = -(\DGradwParGnn \TotalLoss)^{-1} \cdot \GradwTaskWeight \GradwParGnn \TotalLoss$.
Intuitively, IFT allows us to evaluate the $\GradwTaskWeight\ParGnn^*$ locally around
the {approximate best-response} $\ParGnn^*$.
Using the above, we can compute the gradients \GradwTaskWeight\TgtLossOnAuxD as:
\begin{equation}
\begin{aligned}
& \GradwTaskWeight \TgtLossOnAuxD (\ParGnn^*(\TaskWeight)) = 
 \GradwParGnn \TgtLossOnAuxD \cdot \GradwTaskWeight \ParGnn^*(\TaskWeight) \\ 
 & =
- \GradwParGnn \TgtLossOnAuxD \cdot (\DGradwParGnn \TotalLoss)^{-1} \cdot 
\GradwTaskWeight \GradwParGnn \TotalLoss .
\end{aligned}
\end{equation}

We described the entire training process in Algorithm~\ref{alg:1} (Supplementary Section~\ref{sec:app:blo}).
To compute the Hessian inverse and vector products efficiently, we use the iterative
algorithm by Lorraine et al.~\cite{lorraine2020optimizing}, which is summarized in Algorithm \ref{alg:2} (Supplementary Section~\ref{sec:app:blo}).
Intuitively, it uses a Neumann series expansion to approximate the Hessian inverse
with unrolling differentiation for $M$ steps around locally {approximate best-response} $\ParGnn^*$.
%
Following ~\cite{navon2020auxiliary}, in practice, we don't train \ParGnn till convergence
(i.e., $\ParGnn^*$ such that $\GradwParGnn \TotalLoss = 0$).
Instead, we approximate
$\ParGnn^*$ by simultaneously training both \ParGnn and \TaskWeight,
and alternately optimizing \TaskWeight for every $r$ updates of \ParGnn.
%
We refer the readers to ~\cite{lorraine2020optimizing} for theoretical considerations on approximations and convergence.
Note that we use 20\% of the training set as \AuxData instead of using
the validation set to avoid data leakage and unfair comparison with baselines.
Optimizing \TaskWeight on a held-out \AuxData rather than on the training set aligns with the goal
of improving target task generalizability.
\subsection{\BLO with Gradient Rotation (\BLORC)}
\label{sec:methods:blo+rcgrad}
In the previous sections, we discussed \RCGrad, 
which learns to project and scale conflicting gradients using \RWeight,
and \BLO, which learns task weights \TaskWeight but does not explicitly handle gradient conflicts.
In this section, we introduce a novel approach \BLORC that combines the strengths of both \RCGrad and \BLO.
%
Instead of learning the scaling factors \RWeight
by minimizing the combined loss on the training split as in \RCGrad,
\BLORC learns \RWeight that minimizes the target validation loss,
which is similar to the optimization of \TaskWeight in \BLO.
This enables learning \RWeight
that can effectively homogenize conflicting task gradients based on
the generalization performance of the target task.
In \BLORC, the bi-level optimization is employed for learning \RWeight not to balance task losses but
to best align conflicting task gradients.
This addresses the limitation of \BLO in handling gradient conflicts
by incorporating the rotational alignment strategy of \RCGrad.
%
To summarize, \BLORC leverages the learned scaling factors \RWeight via \BLO (Algorithm 1)
to guide the gradient surgery process introduced by \RCGrad (Equation~\ref{equ:rcgrad}).
This dynamically controls the knowledge transfer from auxiliary tasks,
ensuring that the influence of each task is optimally tuned to benefit the target task learning.

\input{tables/sup_cp/result_all}

\section{Experiments}
\label{sec:expts}

\subsection{Experimental Materials}
\label{sec:expts:data}
%
We perform experiments on 8 benchmark classification datasets from MoleculeNet~\cite{wu2018moleculenet}.
We compare our adaptation strategies with simple baselines such as traditional fine-tuning (\FT), 
and vanilla multi-task learning (\MTL)
that assigns equal weights to all auxiliary tasks;
and a more advanced state-of-the-art regularization-based fine-tuning with optimal transport (\GTOT)~\cite{zhang2022fine}.
Additionally, we consider other state-of-the-art gradient surgery-based methods
(\GradSim, \GradScale, \PCGrad) as baselines.
We refer to this group of baselines collectively as \GS methods.
We use the official publicly available checkpoints\footnote{https://github.com/snap-stanford/pretrain-gnns}
of two GNNs:
1) supervised\_contextpred~\cite{hu2019strategies}, denoted as \SUPC,
which is pretrained via self-supervised context prediction and supervised graph-level multi-task learning,
and 2) supervised~\cite{hu2019strategies}, denoted as \SUP,
which is pretrained only via supervised graph-level multi-task learning.
Using such different pretrained GNNs allows a controlled comparison to understand how different pretraining objectives
(with and without self-supervised context prediction task) can influence the adaptation.
Details on auxiliary tasks and datasets are presented in Section~\ref{sec:app:expts} in Supplementary.

\subsection{{Reproducibility and Implementation Details}}
\label{sec:expts:imp}
Following the prior line of research~\cite{hu2019strategies,liu2021pre},
we use scaffold-split for the downstream target tasks, and
use the same atom and bond features as in \GTOT.
All experimental details for the \FT baseline follow the \GTOT fine-tuning setup.
Specifically, we initialized a linear
projection layer on top of the pretrained GNN as the target task classifier.
Across all methods, both the pretrained GNN and task-specific layers are trainable.
For {\FT and adaptation methods}, we train the models for 100 epochs with Adam optimizer
with an initial learning rate $\alpha$ of 0.001,
we use a batch size of \{32, 64, 256\}, an embedding dimension of 300, and a dropout probability of 0.5 for the GNN module.
{For \GTOT experiments, we use the optimal hyper-parameters provided for each dataset,
when finetuned on \SUPC.}
For \MTL experiments, we assign equal weights to all auxiliary tasks.
For \BLO and \BLORC experiments, we use $M=3$ in Algorithm 2,
update \TaskWeight every {$r=\{5,10,20\}$} update of \ParGnn,
and use Adam optimizer with learning rate $\beta$ of 0.001 to update \TaskWeight.
The code is available at
\url{https://github.com/vishaldeyiiest/GraphTA}.

\subsection{Comparison using \SUPC as the pretrained GNN}
\label{sec:expts:supc}

%
Table \ref{tbl:supc:overall} presents an overall comparison when all the auxiliary tasks are used
with \SUPC as the pretrained GNN.
%
Our proposed adaptation strategies, specifically \RCGrad and \BLORC,
outperform all baselines, including other \GS-based adaptation strategies, across all datasets (except ClinTox).
Specifically, compared to the best fine-tuning method, \GTOT, 
\RCGrad demonstrated significant improvement of 2.4\% and 4.8\%
in BACE and BBBP, respectively.
This indicates the efficacy of our proposed rotational alignment in mitigating negative transfer and
improving the generalizability of the pretrained GNN.
Furthermore, \BLORC exhibits significant improvement over fine-tuning methods \FT and \GTOT
in small-scale datasets of as much as 6.3\% and 4.1\%, respectively.
This highlights the efficacy of bi-level optimization
combined with gradient rotation in improving generalizability, especially in limited data regimes.

Additionally,
\RCGrad and \BLORC consistently outperform other gradient surgery-based (\GS) methods.
%
Specifically, compared to \PCGrad,
\RCGrad demonstrates statistically significant improvements in ROC-AUC by 2.5\%, 4.7\%, 0.9\% and 1.0\% 
in ClinTox, BBBP, Tox21, and ToxCast, respectively.
This improvement can be attributed to the rotation component in \RCGrad, 
which not only resolves gradient conflicts but also actively aligns them in a direction favorable to the target task.
Moreover, our proposed methods \RCGrad and \BLORC learn to retain a component of
the conflicting task gradients,
unlike \GradSim which completely discards conflicting gradients.
This ensures that valuable information from auxiliary tasks is not discarded,
thus facilitating more effective knowledge transfer. 

Conversely, \BLO, which learns task weights without explicitly handling gradient conflicts,
performs comparably or slightly worse than \RCGrad, \BLORC, and other \GS-based baselines. 
The suboptimal performance of \BLO, especially in smaller datasets (e.g., SIDER), 
may be attributed to the noisy nature of task gradients,
potentially leading to a poor approximation of hyper-gradients. 
In contrast, \GradScale is more robust to noisy gradients since it adjusts the scale of gradient magnitudes relative to the target task.
Overall, our proposed methods consistently outperform all baselines on smaller datasets (except ClinTox),
while achieving competitive performance on larger ones.

\begin{figure}[!h]
\vspace{-10pt}
    \centering
        \begin{subfigure}{0.23\textwidth}
        \includegraphics[width=\textwidth]{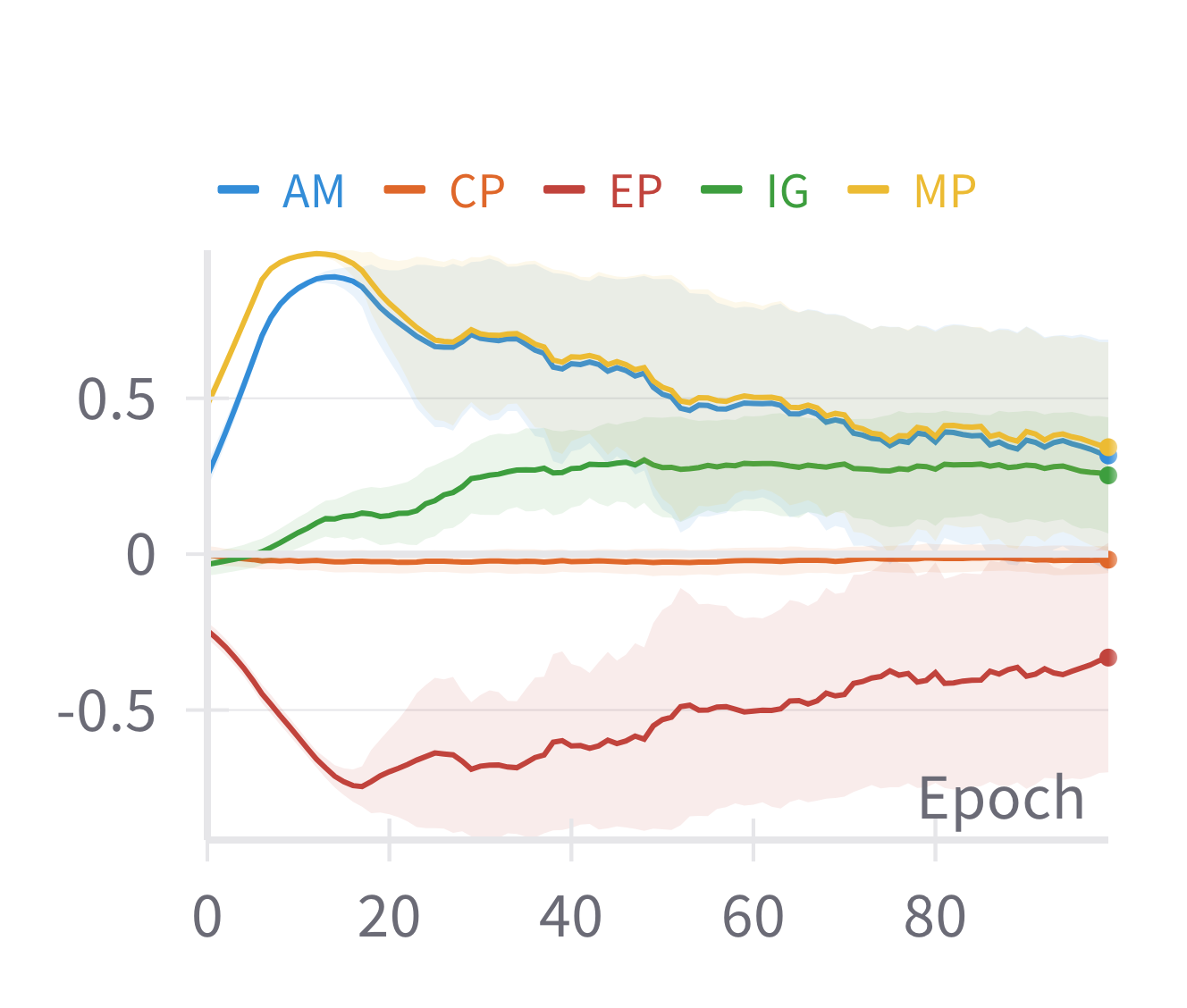}
        \vspace{-15pt}
        \caption{SIDER}
    \end{subfigure}
    \begin{subfigure}{0.23\textwidth}
        \includegraphics[width=\textwidth]{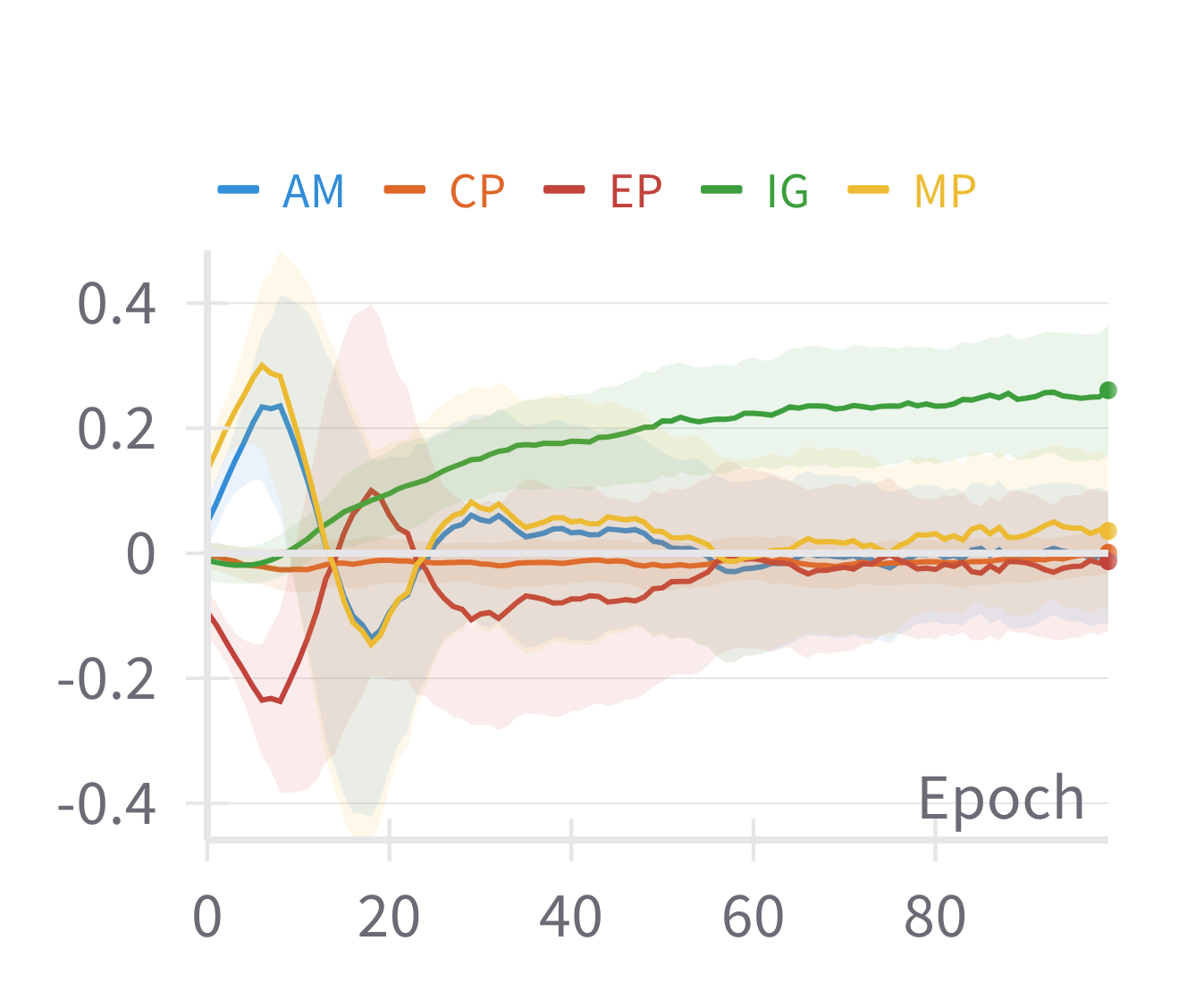}
        \vspace{-15pt}
        \caption{BACE}
    \end{subfigure}
    \begin{subfigure}{0.23\textwidth}
        \includegraphics[width=\textwidth]{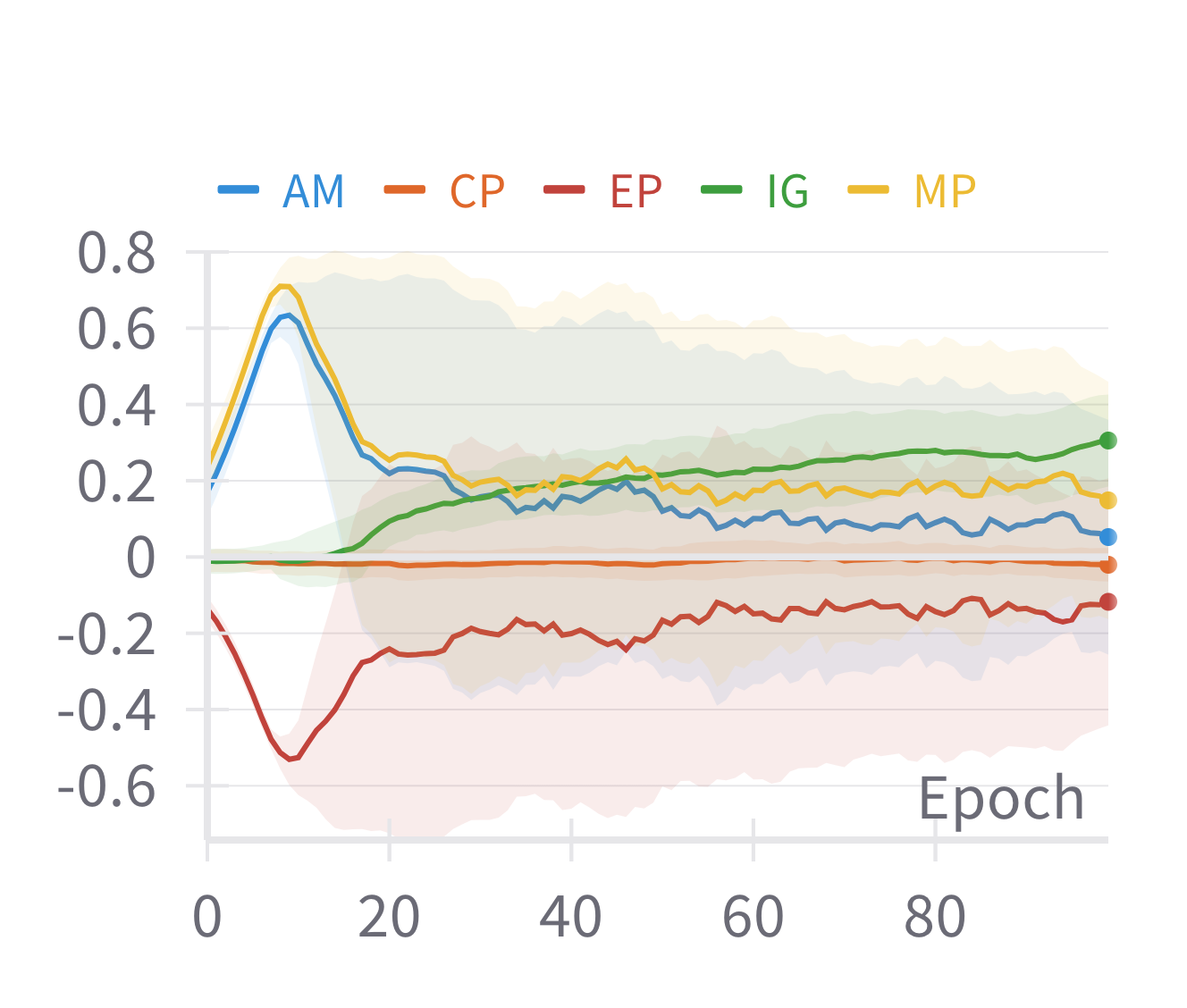}
        \vspace{-15pt}
        \caption{BBBP}
    \end{subfigure}
         \begin{subfigure}{0.23\textwidth}
        \includegraphics[width=\textwidth]{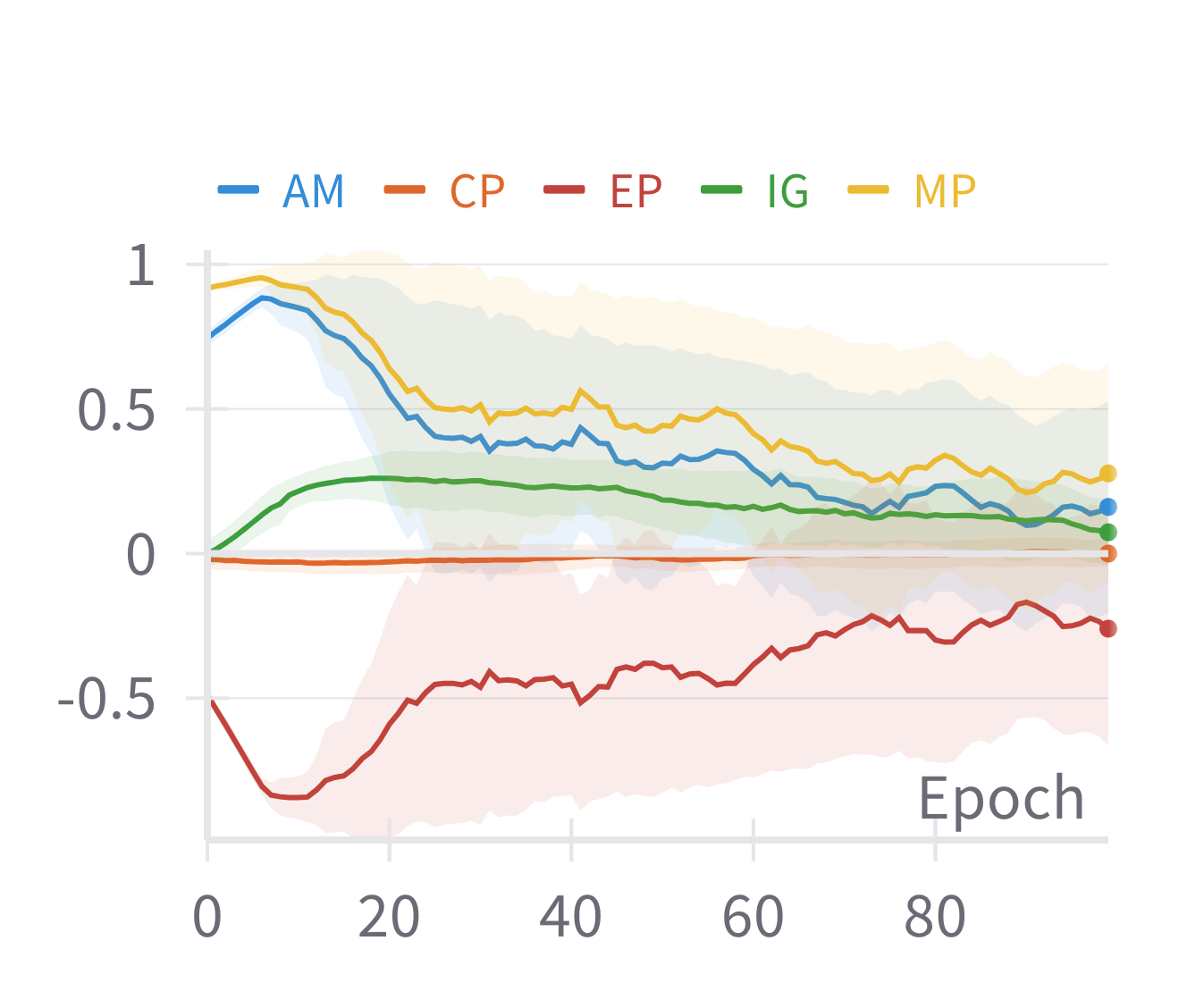}
        \vspace{-15pt}
        \caption{Tox21}
    \end{subfigure}
    \caption{\rebuttal{Target task gradient conflicts with EP and CP tasks.
    \SUPC is adapted with all auxiliary tasks in a \MTL setting.}}
    \label{fig:gsim_mtl}
\end{figure}

In contrast, \MTL, which assigns equal weights to all auxiliary tasks regardless of 
their relevance to the target task,
results in worse performance across all downstream tasks.
Compared to \FT,
\MTL exhibits deteriorations of as much as 9.1\% and 20.6\% in SIDER and ClinTox, respectively.
This indicates that \MTL
leads to drastic negative transfer, where the auxiliary tasks hurt the performance of the target task.
%
On the contrary, all adaptation strategies (including \GS-based baselines) perform better than \MTL
with significant improvements of up to 24.2\%.
%
%
Furthermore, upon analyzing gradient similarities of auxiliary tasks with the target task (Figure~\ref{fig:gsim_mtl}),
we hypothesize that AM, IG, and MP may
benefit the target task better than the other auxiliary tasks.

\input{tables/sup_cp/result_AM+IG+MP}

Table \ref{tbl:supc:overall_sel} presents an overall comparison using only AM, IG, and MP as auxiliary tasks.
Compared to fine-tuning-based methods (\FT and \GTOT), our proposed methods
\RCGrad and \BLORC demonstrate better performance across 6 out of 8 datasets.
Specifically, compared to \GTOT,
\RCGrad achieves significant improvements of 2.6\% and 5.4\% in BACE and BBBP, respectively. 
Furthermore, \RCGrad and \BLORC exhibit better performance than \GS baselines
with significantly improved ROC-AUC of as much as 9.9\% in ClinTox.
Overall, our proposed methods demonstrate significantly improved performance in
smaller datasets compared to fine-tuning and \GS baselines.  
Such consistently superior performance underscores the robustness of our methods, 
particularly in settings where data is limited and the alignment of gradients is crucial.

In contrast with the previous setup, \GS baselines such as
\GradSim and \GradScale exhibit better performance across almost all datasets.
This implies that these methods can be more effective with fewer conflicting tasks, and
may struggle to handle a large number of conflicting tasks (Table~\ref{tbl:supc:overall}).
%
Similarly, with fewer tasks in this setup, \MTL exhibits improved performance compared to the previous setup,
thereby indicating diminished negative transfer. 
This suggests that a smaller and more focused set of 
auxiliary tasks can lead to more efficient and less conflicting learning dynamics.
However, \PCGrad, \RCGrad, and \BLORC,
which partially utilize conflicting gradients, 
show mixed responses to the reduction in the number of auxiliary tasks in this setup.
Specifically, \RCGrad demonstrates improved performance in smaller datasets (except ClinTox)
but a slight decrease in performance 
in larger datasets, compared to their performance in the previous setup.
This can be attributed to the reduced diversity in learning signals provided by a smaller set of auxiliary tasks.
\subsection{Comparison using \SUP as the pretrained GNN}
\label{sec:expts:sup}

\input{tables/sup/result_all}
Table~\ref{tbl:sup:overall} presents an overall comparison of adaptation of \SUP
as the pretrained GNN using all auxiliary tasks.
Similar to our findings in the previous section,
\MTL again results in worse performance compared to fine-tuning methods, thus indicating negative transfer.
On the other hand, our proposed methods, specifically \RCGrad and \BLORC,
demonstrate improved performance over fine-tuning and \GS baselines.
Notably, compared to the best fine-tuning baseline \GTOT,
\BLORC improved ROC-AUC by 6.8\%, 2.2\%, and 4.8\% in
ClinTox, BACE, and BBBP, respectively.
Similarly, compared to the best \GS baseline \GradScale,
\BLORC demonstrates notable improvement of 3.0\%, 11.1\%, and 1.0\%
in SIDER, ClinTox, and BACE, respectively. 
Furthermore, compared to \BLO, which does not explicitly handle conflicting task gradients,
\BLORC yields consistent improvement across most datasets.
Such consistently superior performance of \BLORC implies that
aligning and extracting informative components out of conflicting task gradients
is crucial to improve the generalizablity of pretrained GNNs, regardless of the specific pretraining objective.
%

Following the similar setup of \SUPC experiments with a selected subset of auxiliary tasks,
Table~\ref{tbl:sup:overall_sel} in Supplementary 
presents an overall comparison using \SUP
as the pretrained GNN.
Compared to the previous setup with all auxiliary tasks,
almost all \GS baselines and our proposed method \RCGrad
exhibit improved performance with fewer auxiliary tasks.
This suggests that using a smaller and relevant set of auxiliary tasks
can lead to more efficient adaptation, which holds true across different pretrained GNNs.
Furthermore, compared to the best \GS baseline, \GradScale,
our proposed methods \RCGrad and \BLORC
achieve better or comparable performance, particularly on smaller datasets.
Additionally, \BLORC exhibits significant improvement over \GradSim in Tox21 and ToxCast.
%

%
However, it's worth noting that when using \SUP as the pretrained GNN, all methods, including \RCGrad and \BLORC,
yield slightly worse performance compared to when \SUPC is used as the pretrained GNN.
This observation suggests that the \SUP pretrained GNN
might not capture contextual chemical relationships as effectively as \SUPC, which was pretrained
additionally on the context prediction task.
This subtle difference in performance indicates that the choice of pretrained GNN can have an impact on the overall adaptation process.
{Additional results are presented in
Section~\ref{sec:app:expts} in Supplementary materials}.

\section{Conclusion and Future Work}
\label{sec:conclusion}
In this study, we explored multiple adaptation strategies 
to improve the performance of pretrained GNNs
on downstream molecular property prediction tasks.
To address the poor generalization performance to such diverse downstream tasks,
we introduced two novel methods, \RCGrad and \BLORC,
that learn to align conflicting task gradients.
Our experiments demonstrate that
our proposed methods consistently outperform
all fine-tuning and gradient surgery-based approaches, especially on smaller datasets (except ClinTox).
%
This suggests that the adaptation of pretrained GNNs
can be a promising direction to boost target task performance, especially with limited labeled data.
Our study serves as the first step in exploring the adaptation of pretrained GNNs in molecular property prediction.
In future work, we will explore other adaptation strategies to alleviate noisy gradients
and to improve task selection with sparser task weights.
We will further investigate the benefit of adapting GNNs to diverse downstream molecular regression tasks.

\bibliographystyle{named}
\bibliography{paper}

\appendix

\section{Details on \BLO}
\label{sec:app:blo}
Algorithm~\ref{alg:1} describes the training process of \BLO, and Algorithm~\ref{alg:2} describes the
computation of the gradient \GradwTaskWeight\TgtLossOnAuxD
via approximated Hessian Inverse and vector products.
\begin{figure}[h!]
\begin{minipage}{0.48\textwidth}
\begin{algorithm}[H]
\caption{Learning Task Weights with \BLO}
\label{alg:1}
\begin{algorithmic}[1]
	\State \textbf{Input:} $N$, $r$, $\alpha$
	\State Initialize \TaskWeight with $1/k$,
	 \ParGnn from pretrained GNN, \ParTgt and \ParAux with default Xavier initializer
	\For{$epoch$ from 1 to $N$}
		\State Compute $\TotalLoss = \TgtLoss + \sum_{i=1}^k\TaskWeight_i\iAuxLoss$
		\State $\ParGnn \leftarrow \ParGnn - \alpha\GradwParGnn\TotalLoss$,
		$\ParAux \leftarrow \ParAux - \alpha\GradwParAux\AuxLoss$,
		$\ParTgt \leftarrow \ParTgt - \alpha\GradwParTgt\TgtLoss$
		\If{$epoch \% r$ == 0}
		 	\State $\TaskWeight \leftarrow \TaskWeight - \GradwTaskWeight \TgtLossOnAuxD (\ParGnn(\TaskWeight))$  \Comment{Algorithm \ref{alg:2}}
		\EndIf
	\EndFor
	\State Return \ParGnn, \TaskWeight 
\end{algorithmic}
\end{algorithm}
\end{minipage}
\begin{minipage}{0.48\textwidth}
\begin{algorithm}[H]
\caption{Computing $\GradwTaskWeight \TgtLossOnAuxD (\ParGnn(\TaskWeight))$}
\label{alg:2}
\begin{algorithmic}[1]
	\State \textbf{Input: } \TotalLoss, \TgtLossOnAuxD,
	current $\TaskWeight$, $\ParGnn$ from Algorithm \ref{alg:1},
	$M$, $\beta$
	\State Initialize $p = q = \GradwParGnn\TgtLossOnAuxD|_{\scriptsize{(\TaskWeight, \ParGnn)}}$ ~~~~~
	\Comment{Hessian inverse approximation}
	\For{$j$ from 1 to $M$}
		\State $p = p - \beta p \DGradwParGnn \TotalLoss$
		\State $q = q + p$
	\EndFor
	\State Return $-q\GradwTaskWeight\GradwParGnn \TotalLoss|_{\scriptsize{(\TaskWeight,\ParGnn)}}$
\end{algorithmic}
\end{algorithm}
\end{minipage}
\end{figure}

\section{Experimental Details}
\label{sec:app:expts}

\subsection{On Auxiliary Tasks}
\label{sec:app:expts:aux}

We describe the auxiliary tasks and share key insights behind using them: 
\begin{itemize}
\item Masked Atom Prediction (AM): AM~\cite{hu2019strategies} involves predicting the identity of masked atoms within a molecular graph.
It helps the GNN to learn the local chemical context and relationships between atoms and bonds,
which are crucial for understanding molecular structure and function.
The embedding out of GNN is fed to a linear classifier to predict the atom type of masked atoms.

\item Edge Prediction (EP): EP~\cite{Hamilton:2017tp} focuses on predicting the presence or absence of bonds (edges)
between pairs of atoms in a molecular graph.
It helps the GNN to capture essential local structural information, including connectivity and spatial arrangement of atoms within molecules.
Following existing design\cite{sun2022does}, the dot product of node embeddings is used to predict the existence of a bond.

\item Context Prediction (CP): CP~\cite{hu2019strategies} requires the model to predict neighboring graph structures (context)
based on an anchor structure.
This aids the GNN in distinguishing molecular contexts,
enabling the model to capture subgraph-level information.
The setup of Hu et al.\cite{hu2019strategies} is followed to extract and distinguish positive and negative subgraph contexts.

\item Graph Infomax (IG): IG~\cite{sun2019infograph} maximizes the mutual information between local (node) and global (subgraph) representations.
This helps the GNN to capture structural patterns,
allowing it to understand how atoms form functional groups and larger molecular substructures.
The existing setup~\cite{sun2019infograph} is followed to train a discriminator model that distinguishes between node embeddings
from the same molecular graph and those from a different graph.

\item Motif Prediction (MP): MP~\cite{rong2020self} focuses on predicting the presence of specific recurring substructures (motifs) within a molecule.
It helps the GNN to identify structural motifs indicative of chemical properties or functions.
This task is formulated as a multi-label binary classification problem
with each of 85 motifs\footnote{http://rdkit.org/docs/source/rdkit.Chem.Fragments.html}
extracted from RDKIT~\cite{rdkit} as labels.

\end{itemize}

Each of these tasks focuses on different aspects of molecular graphs, such as local connectivity, spatial arrangement, contextual information, hierarchical organization, and recurring structural patterns. 
In essence, these tasks are designed to equip the model with a richer understanding of molecular structures,
ultimately improving its ability to generalize and make accurate predictions.
Note that designing auxiliary tasks is beyond the scope of this study.

\subsection{Dataset Overview}
\label{sec:app:data}

\input{tables/data}

We perform our adaptation experiments on 8 benchmark classification datasets from MoleculeNet~\cite{wu2018moleculenet}. 
In this section, we give a brief overview and provide preliminary statistics of these datasets.
\begin{itemize}
\item BBBP: measures whether a molecule permeates the blood-brain barrier.
\item BACE: measures whether a molecule inhibit the $\beta$-secretase 1 (BACE-1) enzyme.
\item ClinTox: contains toxicity labels for clinical drugs,
facilitating the assessment of drug safety profiles across various targets.
{It is important to note that these labels reflect both FDA approval outcomes
and clinical trial failures due to toxicity.
Such outcomes are determined by not just the molecular structures of the drugs.
but also by external factors such as genetic predispositions, evaluation methodologies, and environmental conditions.
This complexity can make methodological comparisons challenging.}
\item HIV: measures whether a molecule can prevent antiviral activity against the HIV virus.
\item MUV: compiled and refined from PubChem bioassays, evaluating compound activity across multiple targets.
\item Tox21: measures toxicity across a range of biological pathways used in the 2014 Tox21 challenge.
\item ToxCast: measures compound toxicity across a range of biological systems.
\end{itemize}

\subsection{Additional Figures}
\label{sec:app:res:figs}

\begin{figure}[!hbt]
    \centering
        \begin{subfigure}{0.23\textwidth}
        \includegraphics[width=\textwidth]{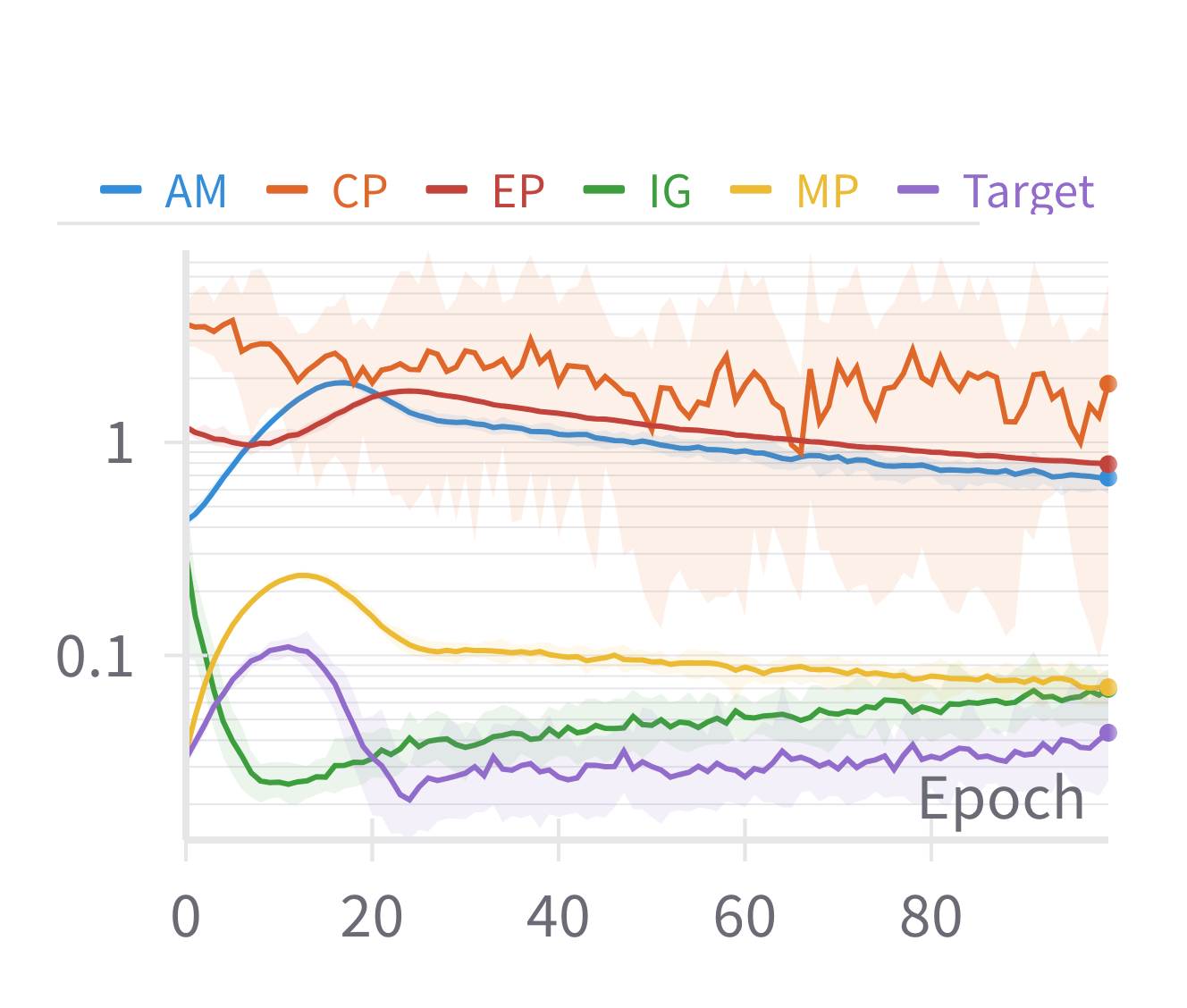}
        \caption{SIDER}
    \end{subfigure}
        \hfill
    \begin{subfigure}{0.23\textwidth}
        \includegraphics[width=\textwidth]{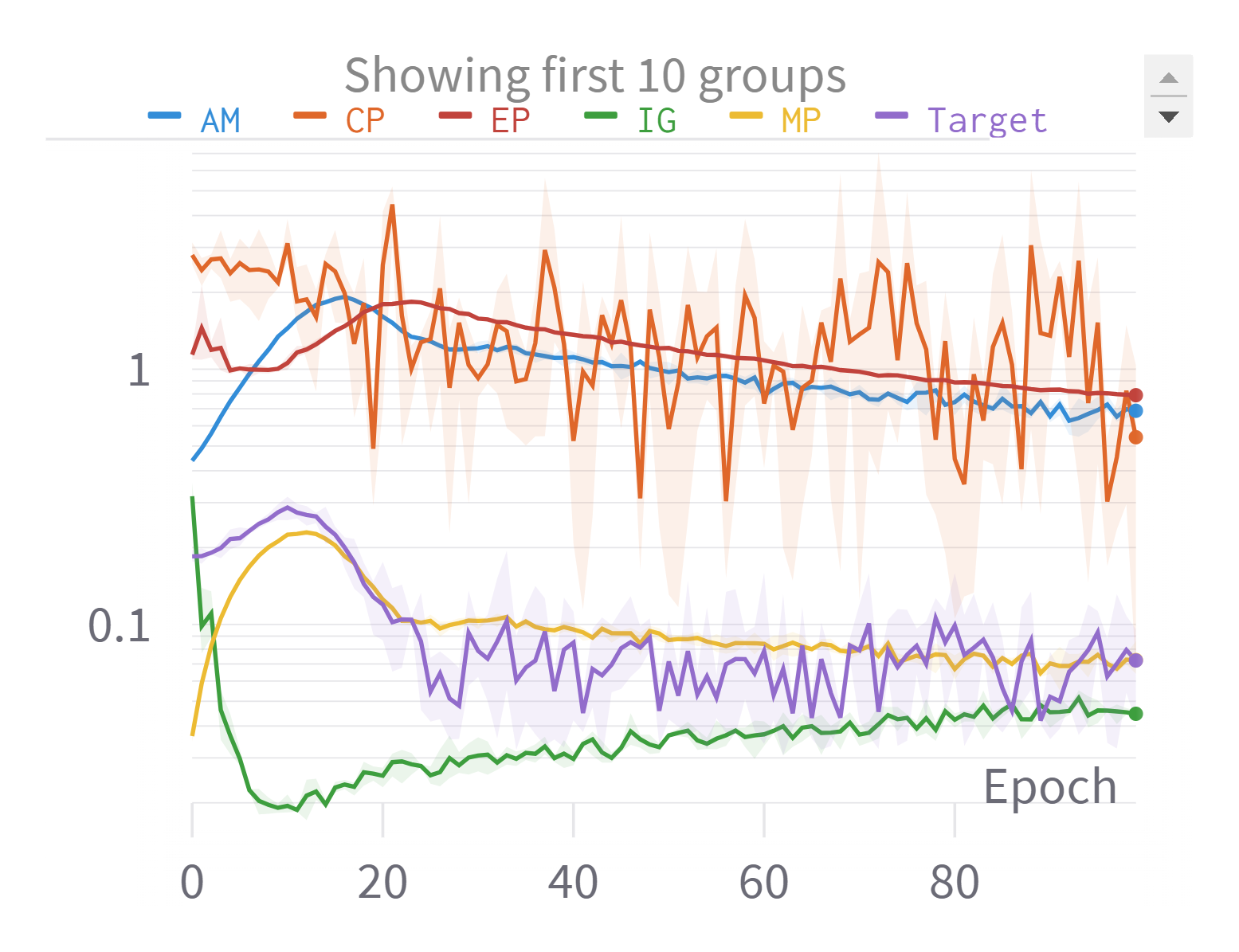}
        \caption{ClinTox}
    \end{subfigure}
    \hfill
    \begin{subfigure}{0.23\textwidth}
        \includegraphics[width=\textwidth]{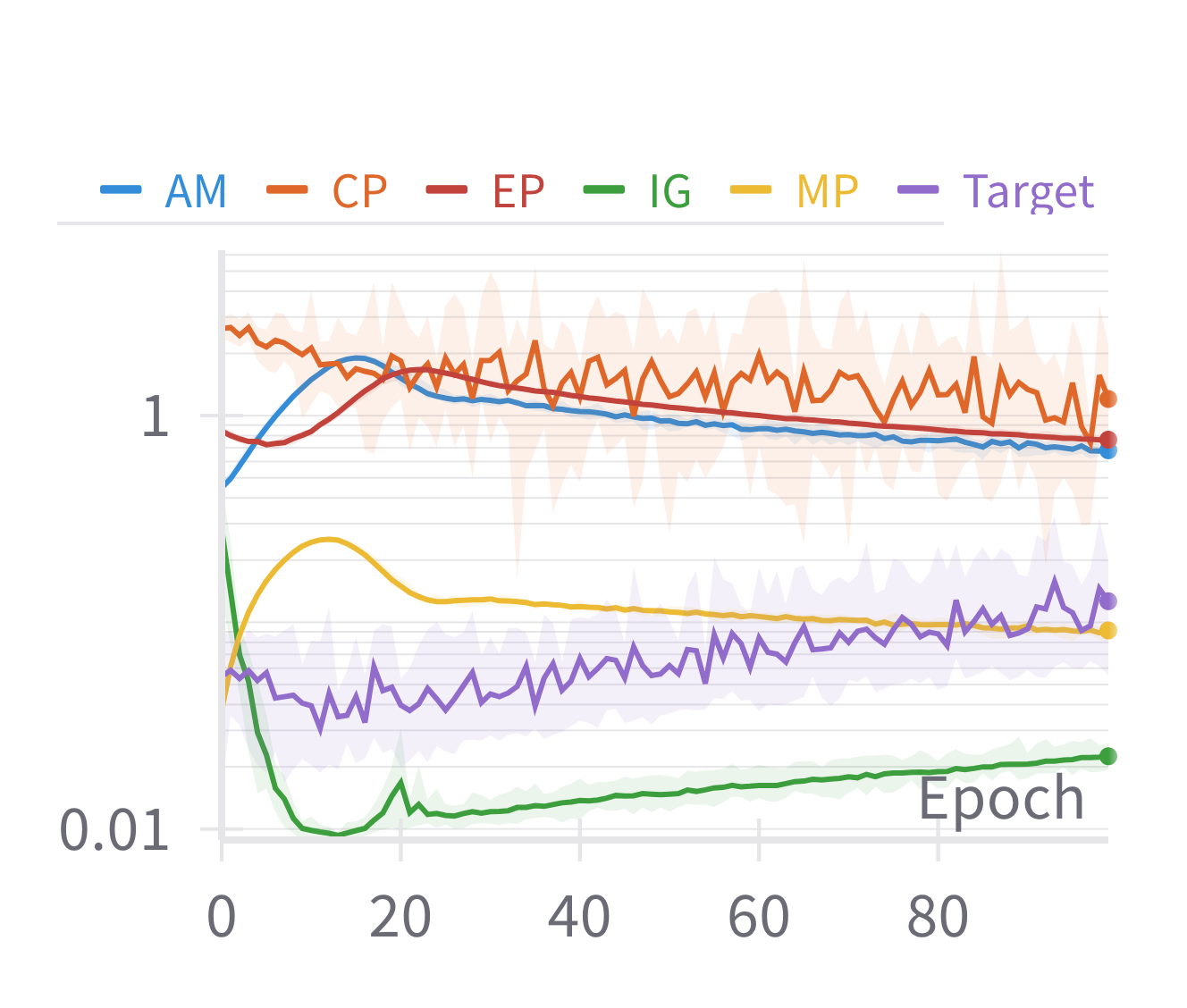}
        \caption{BACE}
    \end{subfigure}
    ~
    \begin{subfigure}{0.23\textwidth}
        \includegraphics[width=\textwidth]{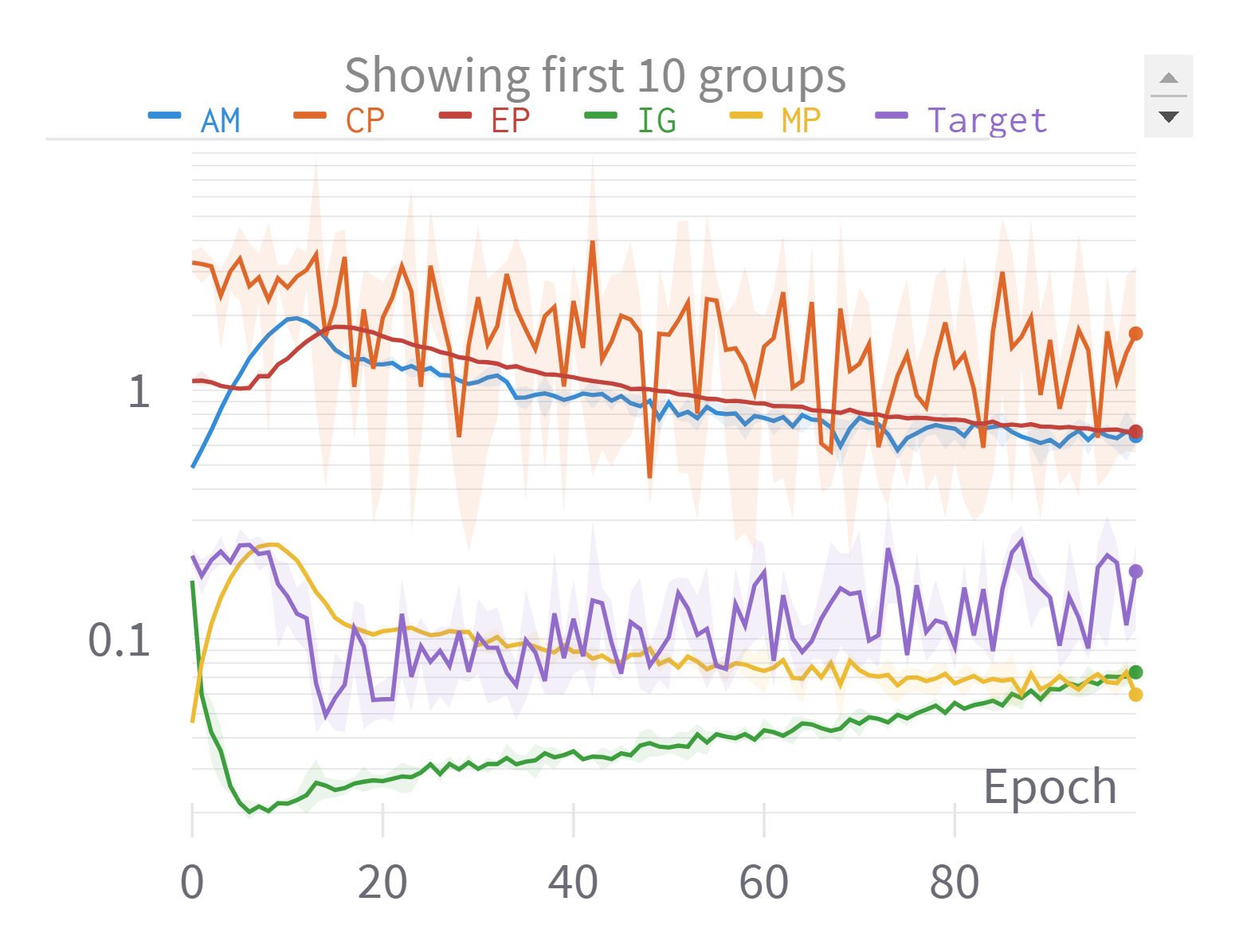}
        \caption{BBBP}
    \end{subfigure}
     \hfill
         \begin{subfigure}{0.23\textwidth}
        \includegraphics[width=\textwidth]{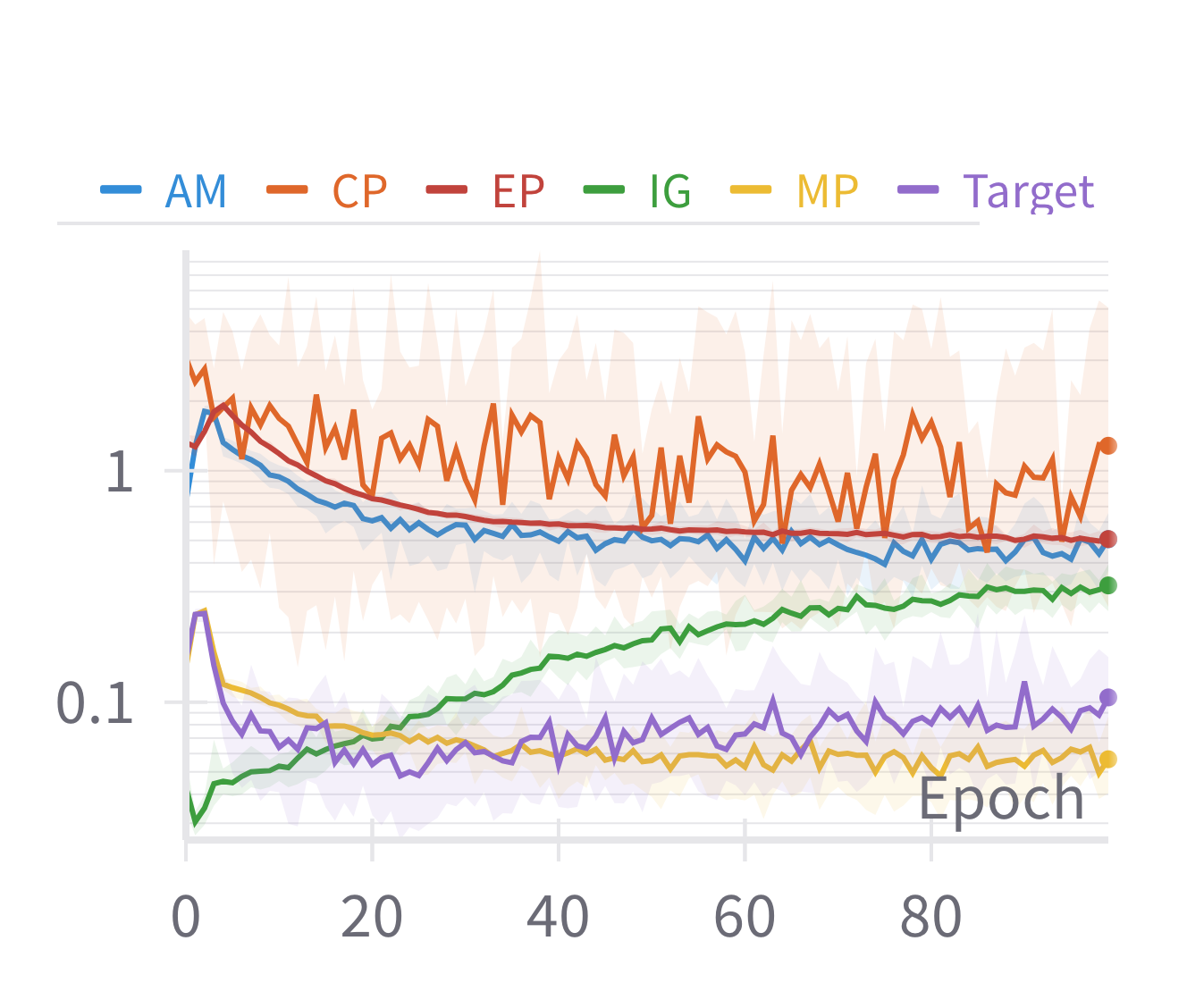}
        \caption{Tox21}
    \end{subfigure}
 \hfill
    \begin{subfigure}{0.23\textwidth}
        \includegraphics[width=\textwidth]{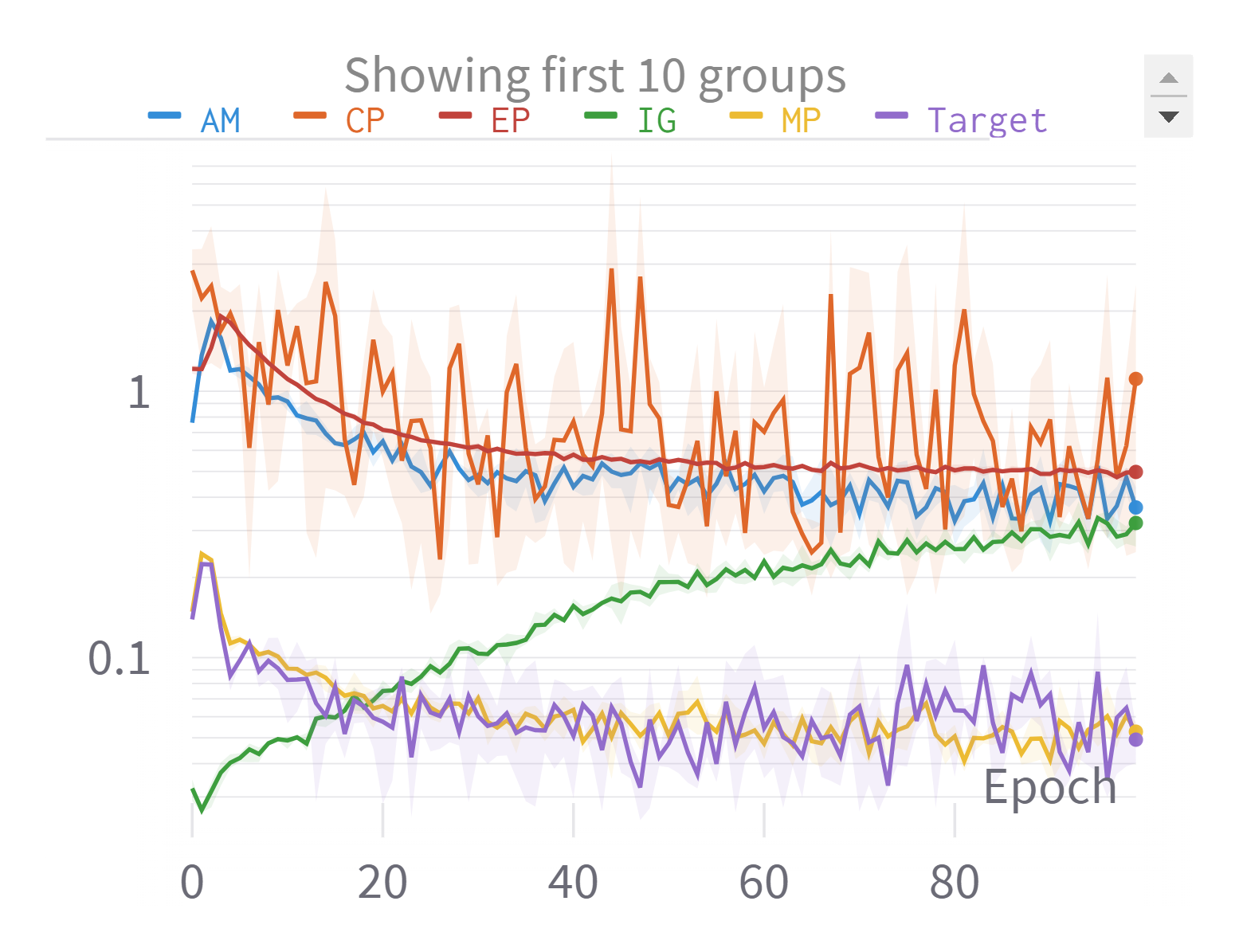}
        \caption{ToxCast}
     \end{subfigure}
    
    \caption{Large variations of scales among task gradients observed across multiple tasks.}
    \label{sfig:supc:gnorm}
\end{figure}

Figure~\ref{sfig:supc:gnorm} demonstrates the varying scales of
auxiliary task gradient magnitudes when \SUPC is adapted using all auxiliary tasks
in a \MTL setting across all datasets.
This indicates the need to adjust the gradient norms as proposed in \GradScale and \RCGrad.
This prevents some auxiliary tasks to dominate over target tasks.

\begin{figure}[!h]
    \centering
        \begin{subfigure}{0.23\textwidth}
        \includegraphics[width=\textwidth]{figs/gsim/sup_cp/sider.png}
        \caption{SIDER}
    \end{subfigure}
        \hfill
    \begin{subfigure}{0.23\textwidth}
        \includegraphics[width=\textwidth]{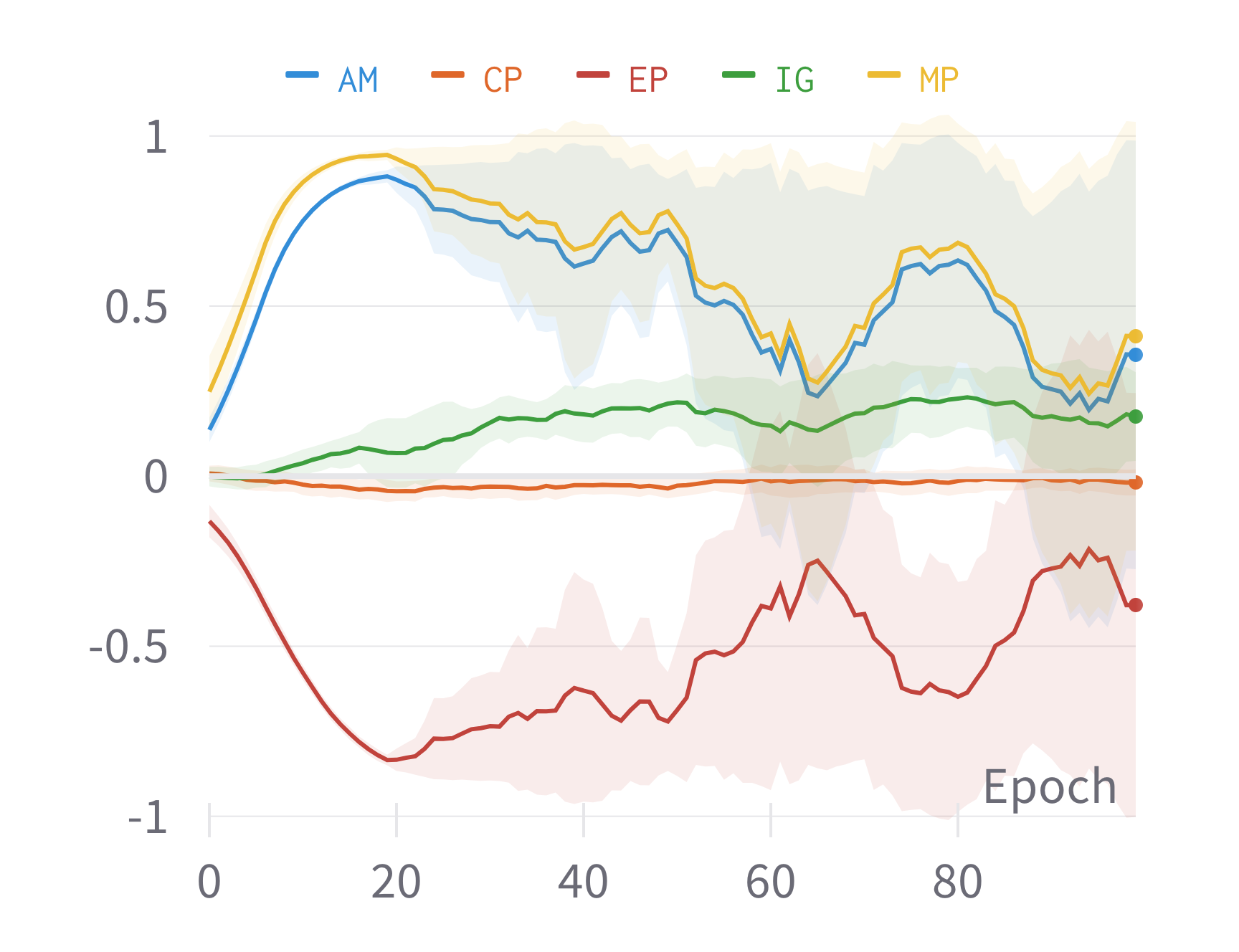}
        \caption{ClinTox}
    \end{subfigure}
    \hfill
    \begin{subfigure}{0.23\textwidth}
        \includegraphics[width=\textwidth]{figs/gsim/sup_cp/bace.png}
        \caption{BACE}
    \end{subfigure}
    ~
    \begin{subfigure}{0.23\textwidth}
        \includegraphics[width=\textwidth]{figs/gsim/sup_cp/bbbp.png}
        \caption{BBBP}
    \end{subfigure}
     \hfill
         \begin{subfigure}{0.23\textwidth}
        \includegraphics[width=\textwidth]{figs/gsim/sup_cp/tox21.png}
        \caption{Tox21}
    \end{subfigure}
 \hfill
    \begin{subfigure}{0.23\textwidth}
        \includegraphics[width=\textwidth]{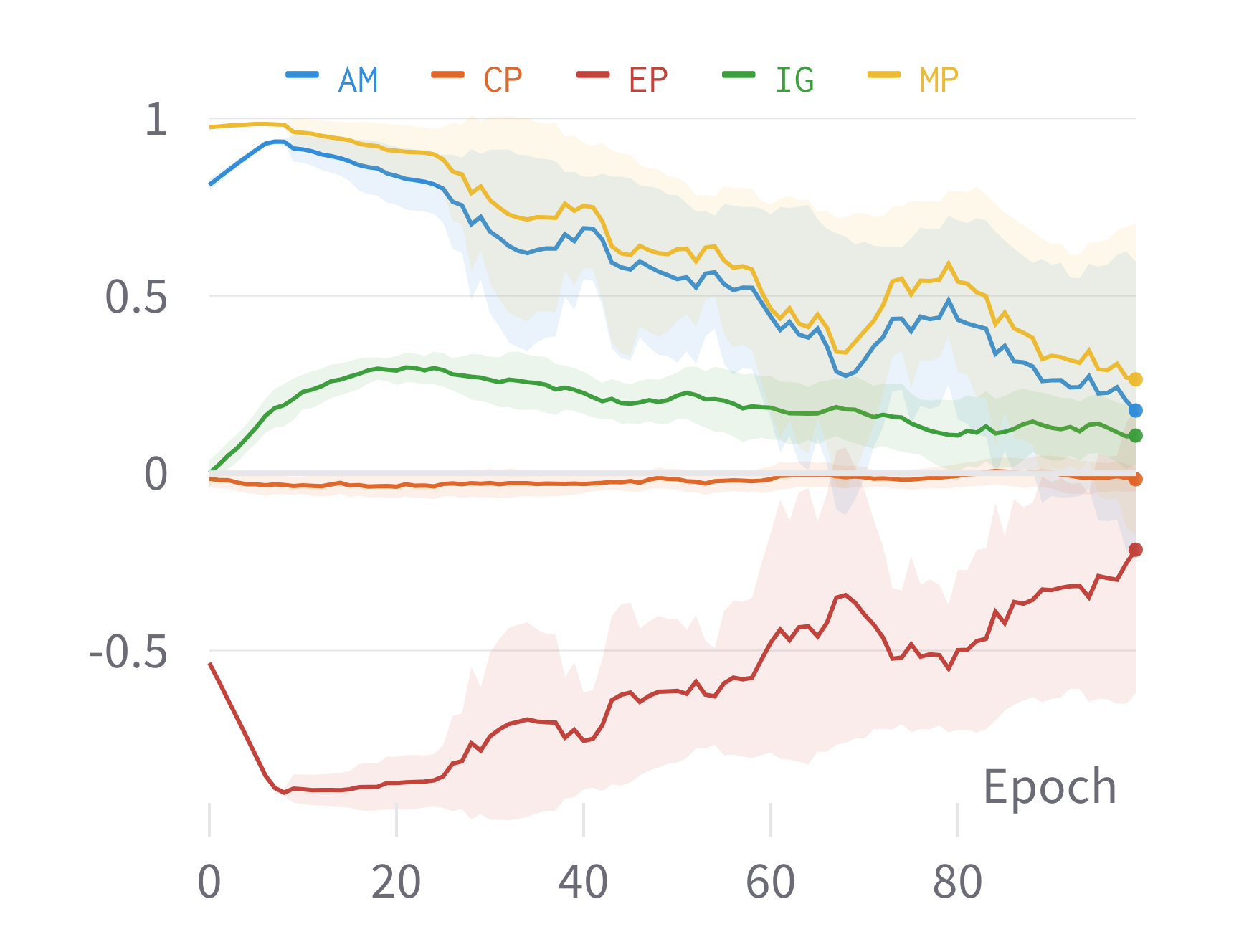}
        \caption{ToxCast}
     \end{subfigure}
    
    \caption{\rebuttal{Target task gradient conflicts with EP and CP tasks.
    \SUPC is adapted with all auxiliary tasks in a \MTL setting.}}
    \label{sfig:supc:gsim_mtl}
\end{figure}

Figure~\ref{sfig:supc:gsim_mtl} demonstrates that
target task gradient conflicts with that of EP and CP tasks across all datasets.
This motivates our experimental comparison of all adaptation strategies
using a smaller set of more relevant auxiliary tasks.

\subsection{Additional Tables}
\label{sec:app:res:tbls}

\input{tables/sup/result_AM+IG+MP}

Table \ref{tbl:sup:overall_sel} presents an overall comparison 
when \SUP is adapted using only AM, IG, and MP as auxiliary tasks.
Compared to fine-tuning-based methods (\FT and \GTOT), our proposed methods
\RCGrad and \BLORC demonstrate better performance across 7 out of 8 datasets.
Specifically, compared to \GTOT,
both \RCGrad and \BLORC achieve significant improvements of up to
5.1\% and 1.8\% in BBBP and ToxCast, respectively. 
Furthermore, \RCGrad and \BLORC exhibit better performance than \GS baselines
with significantly improved ROC-AUC of as much as 7.7\% and 2.2\% in ClinTox and HIV, respectively.
Overall, both \RCGrad and \BLORC outperform fine-tuning methods, while
achieving competitive or better performance than \GS baselines across all datasets.
Such consistently superior performance across multiple setups and pretrained GNNs
underscores the robustness of our methods.

\end{document}

%% file: define.tex
\newcommand{\rebuttal}[1]{#1}  
\newcommand{\bu}[1]{\textbf{\underline{#1}}}

\newcommand{\AuxTask}{\mbox{$\mathop{\mathcal{T}_{a}}\limits$}\xspace}
\newcommand{\iAuxTask}{\mbox{$\mathop{\mathcal{T}_{a,i}}\limits$}\xspace}
\newcommand{\TgtTask}{\mbox{$\mathop{\mathcal{T}_{t}}\limits$}\xspace}

\newcommand{\AuxLoss}{\mbox{$\mathop{\mathcal{L}_{a}}\limits$}\xspace}
\newcommand{\iAuxLoss}{\mbox{$\mathop{\mathcal{L}_{a,i}}\limits$}\xspace}
\newcommand{\TgtLoss}{\mbox{$\mathop{\mathcal{L}_{t}}\limits$}\xspace}
\newcommand{\TgtLossOnAuxD}{\mbox{$\mathop{\mathcal{L}_{t}^{\scriptsize{(\AuxData)}}}\limits$}\xspace}
\newcommand{\TotalLoss}{\mbox{$\mathop{\mathcal{L}_{f}}\limits$}\xspace}

\newcommand{\GradwParGnn}{\mbox{$\mathop{\nabla_{\scriptsize{\ParGnn}}}\limits$}\xspace}
\newcommand{\DGradwParGnn}{\mbox{$\mathop{\nabla^2_{\scriptsize{\ParGnn}}}\limits$}\xspace}
\newcommand{\GradwTaskWeight}{\mbox{$\mathop{\nabla_{\scriptsize{\TaskWeight}}}\limits$}\xspace}

\newcommand{\GradwParTgt}{\mbox{$\mathop{\nabla_{\scriptsize{\ParTgt}}}\limits$}\xspace}
\newcommand{\GradwParAux}{\mbox{$\mathop{\nabla_{\scriptsize{\ParAux}}}\limits$}\xspace}

\newcommand{\iAuxGrad}{\mbox{$\mathop{\mathbf{g}_{a,i}}\limits$}\xspace}
\newcommand{\iAuxGradP}{\mbox{$\mathop{\mathbf{g}^p_{a,i}}\limits$}\xspace}
\newcommand{\iAuxGradR}{\mbox{$\mathop{\mathbf{g}^r_{a,i}}\limits$}\xspace}
\newcommand{\TgtGrad}{\mbox{$\mathop{\mathbf{g}_{t}}\limits$}\xspace}

\newcommand{\SUPC}{\mbox{$\mathop{\mathtt{Sup\text{-}CP}}\limits$}\xspace}
\newcommand{\SUP}{\mbox{$\mathop{\mathtt{Sup}}\limits$}\xspace}

\newcommand{\FT}{\mbox{$\mathop{\mathtt{FT}}\limits$}\xspace}
\newcommand{\MTL}{\mbox{$\mathop{\mathtt{MTL}}\limits$}\xspace}
\newcommand{\GTOT}{\mbox{$\mathop{\mathtt{GTOT}}\limits$}\xspace}
\newcommand{\GS}{\mbox{$\mathop{\mathtt{GS}}\limits$}\xspace}
\newcommand{\GradSim}{\mbox{$\mathop{\mathtt{GCS}}\limits$}\xspace}
\newcommand{\GradScale}{\mbox{$\mathop{\mathtt{GNS}}\limits$}\xspace}
\newcommand{\BLO}{\mbox{$\mathop{\mathtt{BLO}}\limits$}\xspace}
\newcommand{\PCGrad}{\mbox{$\mathop{\mathtt{PCGrad}}\limits$}\xspace}
\newcommand{\RCGrad}{\mbox{$\mathop{\mathtt{RCGrad}}\limits$}\xspace}
\newcommand{\BLORC}{\mbox{$\mathop{\mathtt{BLO}\text{+}\mathtt{RCGrad}}\limits$}\xspace}

\newcommand{\ParGnn}{\mbox{$\Theta$}\xspace}
\newcommand{\ParTgt}{\mbox{$\Psi$}\xspace}
\newcommand{\ParAux}{\mbox{$\Phi$}\xspace}

\newcommand{\TaskWeight}{\mbox{$\mathbf{w}$}\xspace}
\newcommand{\RWeight}{\mbox{$\mathbf{s}$}\xspace}

\newcommand{\AuxData}{\mbox{$\mathop{\mathcal{A}}\limits$}\xspace}

%% file: tables/sup_cp/result_all.tex
\begin{table*}[hbt!]
\centering
\caption{{Test ROC-AUC using $\AuxTask$=\{AM,CP,EP,IG,MP\}
and \SUPC}}
\label{tbl:supc:overall}
\begin{small}
  \begin{threeparttable}
      \begin{tabular}{
          @{\hspace{5pt}}l@{\hspace{5pt}}
          @{\hspace{5pt}}r@{\hspace{5pt}}
          @{\hspace{5pt}}r@{\hspace{5pt}}
          @{\hspace{5pt}}r@{\hspace{5pt}}
          @{\hspace{5pt}}r@{\hspace{8pt}}
          @{\hspace{5pt}}r@{\hspace{5pt}}
          @{\hspace{5pt}}r@{\hspace{5pt}}
          @{\hspace{5pt}}r@{\hspace{5pt}}
          @{\hspace{5pt}}r@{\hspace{5pt}}
          }
          \toprule
	 Method & SIDER & ClinTox & BACE & BBBP &	Tox21& ToxCast & HIV & MUV \\
	 \midrule
	 \FT & 61.82 (0.53) & \bu{71.10} (1.40) & 82.86 (0.87) & 67.57 (1.39) 
	 & 77.05 (0.34) & 66.02 (0.18) & 78.70 (0.80) & 80.64 (0.51)
	 \\
	 \GTOT & 62.24 (0.34) & {70.03} (1.58) & 83.67 (1.75) & 69.01 (1.95) 
	 & \textbf{77.08} (0.66) & 65.45 (0.45) & \bu{80.05} (0.57) & 82.09 (3.13) 
	 \\
	 \hline
	 \MTL & 56.22 (0.82) & 56.41 (3.43) & 80.04 (1.48) & 64.88 (1.23) 
	 & 74.42 (0.34) & 64.53 (0.38) & 76.79 (0.26) & 81.68 (0.49) 
	 \\
	 \hline
	 \GradSim & 59.94 (0.53) & 62.77 (2.17) & 85.60 (0.63) & 71.16 (0.52) 
	 & 74.76 (0.42) & 66.05 (0.19) & 76.94 (1.30) & 76.65 (1.17) 
	 \\
	 \GradScale & {62.48} (0.56) & 67.94 (1.02) & 84.80 (0.34) 
	 & 70.94 (0.86) & 76.44 (0.24) & 66.19 (0.21) & 78.23 (0.44) & \textbf{83.50} (1.14) 
	 \\
	\PCGrad & 62.09 (0.62) & 67.60 (1.88) & 84.42 (1.23) & 69.14 (1.25) 
	& 76.58 (0.77) & 65.81 (0.61) & 77.76 (1.08) & 78.62 (0.29)
	\\
	 \hline
	 \BLO & 60.70 (2.37) & 68.29 (3.02) & 85.14 (1.23)\rlap{*} & 69.80 (0.68) 
	 & 76.57 (0.44) & 65.80 (0.79) & 79.16 (0.47) & 82.19 (1.40)
	 \\
	 \RCGrad & \textbf{62.49} (0.65) & \textbf{70.07} (1.70)\rlap{$\dagger$} & \textbf{85.65} (0.60)\rlap{*} & \bu{72.35} (1.28)\rlap{*$\dagger$} 
	 & \bu{77.26} (0.38)\rlap{$\dagger$} & \bu{66.49} (0.30)\rlap{*$\dagger$} & \textbf{79.39} (0.63) & {83.07} (1.26)
	 \\
	 \BLORC & \bu{62.94} (0.66)\rlap{*} & 69.59 (2.12)\rlap{$\dagger$} & \bu{86.10} (0.35)\rlap{*} & \textbf{71.81} (1.57)\rlap{*} 
	 & 76.62 (0.29) & \textbf{66.38} (0.29)\rlap{*} & 79.03 (1.12) & \bu{83.92} (1.03) 
	 \\
	 \bottomrule
\end{tabular}

\begin{tablenotes}
	\footnotesize
        \setlength\labelsep{0pt}
		\item We report the mean (and standard deviation) over {10} different seeds with scaffold splitting.
Best- and second best-performing models are in \bu{bold} and \textbf{bold}. 
Tasks are presented in increasing order of size. 
{* and $\dagger$ indicate statistical significance compared to the best finetuning and \GS baselines, respectively.
Statistical significance is determined based on the Wilcoxon signed rank test with $p<0.05$.}
		\par
	\end{tablenotes}

\end{threeparttable}
\end{small}
\end{table*}

%% file: tables/sup_cp/result_AM+IG+MP.tex
\begin{table*}[hbt!]
\centering
\caption{{Test ROC-AUC using $\AuxTask$=\{\text{AM,IG,MP}\} and \SUPC}}
\label{tbl:supc:overall_sel}
\begin{small}
  \begin{threeparttable}
      \begin{tabular}{
          @{\hspace{5pt}}l@{\hspace{5pt}}
          @{\hspace{5pt}}r@{\hspace{5pt}}
          @{\hspace{5pt}}r@{\hspace{5pt}}
          @{\hspace{5pt}}r@{\hspace{5pt}}
          @{\hspace{5pt}}r@{\hspace{5pt}}
          @{\hspace{5pt}}r@{\hspace{5pt}}
          @{\hspace{5pt}}r@{\hspace{5pt}}
          @{\hspace{5pt}}r@{\hspace{5pt}}
          @{\hspace{5pt}}r@{\hspace{5pt}}
          }
          \toprule
	 Method & SIDER & ClinTox & BACE & BBBP &	Tox21& ToxCast & HIV & MUV \\
	 \midrule
	 \FT & 61.82 (0.53) & \bu{71.10} (1.40) & 82.86 (0.87) & 67.57 (1.39) 
	 & 77.05 (0.34) & 66.02 (0.18) & 78.70 (0.80) & 80.64 (0.51) 
	 \\
	 \GTOT & 62.24 (0.34) & {70.03} (1.58) & 83.67 (1.75) & 69.01 (1.95) 
	 & \textbf{77.08} (0.66) & 65.45 (0.45) & \bu{80.05} (0.57) & 82.09 (3.13) 
	 \\
	 \hline
	 \MTL & 59.15 (1.84) & 62.01 (1.87) & 83.60 (0.43) & 71.67 (4.44) 
	 & 75.64 (0.37) & 65.14 (0.21) & 78.18 (1.07) & 81.26 (1.90) 
	 \\
	 \hline
	 \GradSim & {62.83} (0.70) & 64.62 (1.83) & 84.17 (0.87) & 70.49 (4.26) 
	 & \bu{77.35} (0.20) & 66.03 (0.12) & 77.59 (1.24) & 80.17 (3.26) 
	 \\
	 \GradScale & 62.62 (0.49) & 63.42 (2.19) & 84.29 (0.97) & 71.79 (3.72) 
	 & 76.50 (0.39) & 66.12 (0.20) & 78.25 (0.60) & {82.42} (0.47) 
	 \\
	 \PCGrad & 61.42 (1.69) & 63.44 (2.90) & 83.92 (1.23) & 70.86 (4.54) 
	 & 76.73 (0.89) & 65.96 (0.71) & 77.38 (1.10) & 80.45 (2.34) 
	 \\
	 \hline
	 \BLO & \textbf{62.85} (0.77) & 67.52 (3.27)\rlap{$\dagger$} & \textbf{84.79} (0.62) & {71.93} (3.19)\rlap{*} 
	 & 76.88 (0.26) & \textbf{66.29} (0.26)\rlap{*} & 79.21 (0.33) & 81.86 (1.19) 
	 \\
	 \RCGrad & \bu{63.12} (0.38)\rlap{*$\dagger$} & 69.91 (1.22)\rlap{$\dagger$} & \bu{85.86} (0.38)\rlap{*$\dagger$} & \bu{72.76} (1.05)\rlap{*} 
	 & 76.86 (0.38) & \bu{66.37} (0.16)\rlap{*$\dagger$} & {79.17} (0.32) & \bu{82.68} (1.92) 
	 \\
	 \BLORC & 62.44 (0.28) & \textbf{70.99} (2.31)\rlap{$\dagger$} & {84.75} (0.53) & \textbf{72.03} (3.83)\rlap{*}
	 & 76.64 (0.28) & {66.25} (0.22)\rlap{*$\dagger$} & \textbf{79.75} (0.81) & \textbf{82.65} (3.36) 
	 \\	
	 \bottomrule
\end{tabular}

\begin{tablenotes}
	\footnotesize
        \setlength\labelsep{0pt}
		\item 
Best- and second best-performing models are in \textbf{\underline{bold}} and \textbf{bold}. 
{* and $\dagger$ indicate statistical significance compared to the best baselines
based on the Wilcoxon signed rank test with $p < 0.05$.}
		\par
	\end{tablenotes}

\end{threeparttable}
\end{small}
\end{table*}

%% file: tables/sup/result_all.tex
\begin{table*}[!hbt]
\centering
\caption{{Test ROC-AUC using $\AuxTask$=\{AM,CP,EP,IG,MP\} and \SUP}}
\label{tbl:sup:overall}
\begin{small}
  \begin{threeparttable}
      \begin{tabular}{
          @{\hspace{5pt}}l@{\hspace{5pt}}
          @{\hspace{5pt}}r@{\hspace{5pt}}
          @{\hspace{5pt}}r@{\hspace{5pt}}
          @{\hspace{5pt}}r@{\hspace{5pt}}
          @{\hspace{5pt}}r@{\hspace{5pt}}
          @{\hspace{5pt}}r@{\hspace{5pt}}
          @{\hspace{5pt}}r@{\hspace{5pt}}
          @{\hspace{5pt}}r@{\hspace{5pt}}
          @{\hspace{5pt}}r@{\hspace{5pt}}
          }
          \toprule
	 Method & SIDER & ClinTox & BACE & BBBP &	Tox21& ToxCast & HIV & MUV \\
	 \midrule
	 \FT & 61.85 (0.68) & 54.16 (5.25) & 75.76 (0.65) & 66.34 (0.82) 
	 & 75.64 (0.22) & 63.52 (0.23) & 72.84 (0.85) & \bu{80.46} (0.19) \\
	 \GTOT & \textbf{62.38} (0.39) & 55.64 (7.49) & 75.82 (2.10) & 66.26 (1.87)
	  & 75.25 (1.11) & 64.00 (0.55) & 74.93 (1.50) & \textbf{80.42} (0.42)  \\
	 \hline
	 \MTL & 55.18 (0.96) & 47.33 (1.84) & 64.84 (2.43) & 63.62 (1.08) 
	 & 73.15 (0.44) & 62.06 (2.00)  & 63.25 (5.15) & 69.21 (8.51)\\
	 \hline
	 \GradSim & 58.39 (0.59) & 50.05 (1.48) & 74.59 (0.61) & 66.67 (2.41) 
	 & 74.36 (0.43) & 63.94 (0.35) & 72.23 (0.24) & 62.99 (5.35) \\
	 \GradScale & 60.57 (2.04) & 53.52 (5.44) & 76.69 (0.88) & 68.67 (0.42) 
	 & 75.37 (0.34) & 63.49 (0.12) & 74.41 (0.19) & 79.72 (0.17) \\
	 \PCGrad & 59.83 (0.53) & 53.07 (5.12) & 71.17 (6.65) & 67.18 (1.12) 
	 & 74.26 (0.53) & 63.95 (0.42) & 71.80 (0.45) & 79.31 (0.74)\\
	 \hline
	 \BLO & 60.65 (2.66) & 56.10 (4.77) & 75.11 (1.19) & 67.81 (1.09)\rlap{*} 
	 & 74.57 (0.59) & \textbf{64.20} (0.44) & 75.05 (0.74)\rlap{$\dagger$} & 78.12 (0.68) \\ 
	 \RCGrad & 61.38 (0.74) & \textbf{57.36} (3.75) & \textbf{77.00} (1.03) & \textbf{68.73} (0.76)\rlap{*} 
	 & \textbf{75.67} (0.49) & 63.91 (0.23) & \textbf{75.60} (0.26)\rlap{$\dagger$} & 79.37 (1.74)\\
	 \BLORC & \bu{62.41} (0.81) & \bu{59.45} (3.33)\rlap{$\dagger$} & \bu{77.47} (0.79) & \bu{69.45} (0.70)\rlap{*} 
	 & \bu{76.08} (0.34)\rlap{$\dagger$} & \bu{64.60} (0.28)\rlap{*$\dagger$} & \bu{75.80} (0.41)\rlap{$\dagger$} & 79.97 (1.11)  \\
	 \bottomrule
\end{tabular}

\begin{tablenotes}
	\footnotesize
        \setlength\labelsep{0pt}
		\item 
Best- and second best-performing models are in \bu and \textbf{bold}. 
* and $\dagger$ indicate statistical significance compared to the best baselines
based on the Wilcoxon signed rank test with $p < 0.05$.
		\par
	\end{tablenotes}

\end{threeparttable}
\end{small}
\end{table*}

%% file: tables/data.tex
\begin{table*}[!h]
\centering
\caption{Overview of benchmark molecular property prediction datasets}
\label{tbl:data}
  \begin{threeparttable}
      \begin{tabular}{
          @{\hspace{3pt}}l@{\hspace{3pt}}
          @{\hspace{3pt}}r@{\hspace{3pt}}
          @{\hspace{3pt}}r@{\hspace{3pt}}
          @{\hspace{3pt}}r@{\hspace{3pt}}
          @{\hspace{3pt}}r@{\hspace{3pt}}
          @{\hspace{3pt}}r@{\hspace{3pt}}
          @{\hspace{3pt}}r@{\hspace{3pt}}
          @{\hspace{3pt}}r@{\hspace{3pt}}
          @{\hspace{3pt}}r@{\hspace{3pt}}
          }
          \toprule
	 Dataset & BBBP & Tox21 & ToxCast & SIDER & ClinTox & MUV & HIV & BACE \\
	 \midrule
	 No. mols 		& 2,039 & 7.831 & 8,575 & 1,427 & 1,478 & 93,087 & 41,127 & 1,513 \\
	 No. tasks		& 1 & 12 & 617 & 27 & 2 & 17 & 1 & 1 \\
	 Avg. atoms 	& 24.06  & 18.57 & 18.78 & 33.64 & 26.16 & 24.23 & 25.51 & 34.09 \\
	 Avg. diameter	& 11.32 & 9.62 & 9.49 & 14.14 & 12.39 & 12.79 & 11.98 & 15.22 \\

	 \bottomrule
	 \end{tabular}

\end{threeparttable}
\end{table*}

%% file: tables/sup/result_AM+IG+MP.tex
\begin{table*}[h!]
\centering
\caption{{Test ROC-AUC using $\AuxTask$=\{\text{AM,IG,MP}\} and \SUP}}
\label{tbl:sup:overall_sel}
\begin{small}
  \begin{threeparttable}
      \begin{tabular}{
          @{\hspace{5pt}}l@{\hspace{5pt}}
          @{\hspace{5pt}}r@{\hspace{5pt}}
          @{\hspace{5pt}}r@{\hspace{5pt}}
          @{\hspace{5pt}}r@{\hspace{5pt}}
          @{\hspace{5pt}}r@{\hspace{5pt}}
          @{\hspace{5pt}}r@{\hspace{5pt}}
          @{\hspace{5pt}}r@{\hspace{5pt}}
          @{\hspace{5pt}}r@{\hspace{5pt}}
          @{\hspace{5pt}}r@{\hspace{5pt}}
          }
          \toprule
	 Method & SIDER & ClinTox & BACE & BBBP &	Tox21& ToxCast & HIV & MUV \\
	 \midrule
	 \FT & 61.85 (0.68) & 54.16 (5.25) & 75.76 (0.65) & 66.34 (0.82) 
	 & \textbf{75.64} (0.22) & 63.52 (0.23) & 72.84 (0.85) & \bu{80.46} (0.19) \\
	 \GTOT & {62.38} (0.39) & 55.64 (7.49) & 75.82 (2.10) & 66.26 (1.87)
	  & 75.25 (1.11) & 64.00 (0.55) & 74.93 (1.50) & \textbf{80.42} (0.42)  \\
	 \hline
	 \MTL & 56.24 (2.79) & 53.25 (2.60) & 75.92 (1.06) & 68.72 (0.73) 
	 & 72.22 (0.62) & 62.94 (0.32) & 71.84 (0.95) & 74.81 (0.48) \\
	 \hline
	 \GradSim & 61.31 (0.65) & 50.22 (1.60) & 75.54 (1.12) & 65.23 (1.89) 
	 & 75.01 (0.30) & 64.45 (0.27) & 74.03 (0.52) & 75.20 (1.99) \\
	 \GradScale & \textbf{62.47} (0.49) & 55.08 (4.51) & \textbf{77.28} (1.38) & 69.55 (1.10) 
	 & 74.95 (0.41) & 63.94 (0.23) & 74.13 (0.34) & 77.05 (1.53)\\
	 \PCGrad & 57.66 (2.41) & 52.20 (3.22) & 76.55 (0.96) & 69.11 (0.62) 
	 & 73.01 (0.90) & 63.59 (0.48) & 71.88 (1.15) & 75.28 (1.32) \\
	 \hline
	 \BLO & 61.70 (0.86) & 56.79 (3.67) & 75.25 (1.54) & 68.00 (0.88)\rlap{*} 
	 & 74.53 (0.33) & 64.44 (0.73) & 75.15 (0.35)\rlap{$\dagger$} & 76.97 (3.12) \\
	 \RCGrad & 62.10 (1.04) & \textbf{58.64} (1.66)\rlap{$\dagger$} & \bu{77.64} (0.80) & \textbf{69.63} (0.84)\rlap{*} 
	 & 75.08 (0.53) & \textbf{65.09} (0.37)\rlap{*$\dagger$} & \textbf{75.63} (0.24)\rlap{$\dagger$}  &  78.08 (2.78)\\
	 \BLORC & \bu{62.55} (0.85) & \bu{59.31} (3.59)\rlap{$\dagger$} & 77.22 (1.53) & \bu{69.67} (0.93)\rlap{*} 
	 & \bu{75.74} (0.58)\rlap{$\dagger$} & \bu{65.18} (0.44)\rlap{*$\dagger$} & \bu{75.78} (0.21)\rlap{$\dagger$} & 78.37 (2.31) \\
	 \bottomrule
\end{tabular}

\begin{tablenotes}
	\footnotesize
        \setlength\labelsep{0pt}
		\item 
Best- and second best-performing models are in \textbf{\underline{bold}} and \textbf{bold}. 
* and $\dagger$ indicate statistical significance compared to the best baselines
based on the Wilcoxon signed rank test with $p < 0.05$.\
		\par
	\end{tablenotes}

\end{threeparttable}
\end{small}
\end{table*}